\documentclass{article}

\usepackage{arxiv}

\usepackage[utf8]{inputenc} 
\usepackage[T1]{fontenc}    
\usepackage{hyperref}       
\usepackage{url}            
\usepackage{booktabs}       
\usepackage{amsfonts}       
\usepackage{nicefrac}       
\usepackage{microtype}      
\usepackage{lipsum}		
\usepackage{graphicx}
\usepackage[square, comma, sort&compress, numbers]{natbib}
\usepackage{doi}

\usepackage{microtype}
\usepackage{graphicx}
\usepackage{subfigure}
\usepackage{booktabs} 
\usepackage{color}
\usepackage[absolute]{textpos}

\usepackage{hyperref}

\usepackage{wrapfig}
\usepackage{graphicx}
\usepackage{float}
\usepackage{subfigure}
\usepackage{authblk}
\newcommand*{\affaddr}[1]{#1} 

\usepackage{amsmath}
\usepackage{extarrows}   
\usepackage{chemarrow}   

\title{Grid-SD2E: A General Grid-Feedback in a System for Cognitive Learning}

\author{%
\textbf{
Jingyi Feng and Chenming Zhang
}\\
\affaddr{Institute of Artificial Intelligence, School of Computer Science, Wuhan University}
\thanks{Email: fjy2035@gmail.com (Jingyi Feng)} \\
}

\date{}

\renewcommand{\headeright}{}
\renewcommand{\undertitle}{}
\renewcommand{\shorttitle}{Grid-SD2E: A General Grid-Feedback in a System for Cognitive Learning (Neural Decoding)}

\hypersetup{
pdftitle={A template for the arxiv style},
pdfsubject={q-bio.NC, q-bio.QM},
pdfauthor={David S.~Hippocampus, Elias D.~Striatum},
pdfkeywords={First keyword, Second keyword, More},
}

\begin{document}
\maketitle

\begin{abstract}
\label{abstract}
Comprehending how the brain interacts with the external world through generated neural data is crucial for determining its working mechanism, treating brain diseases, and understanding intelligence. 
Although many theoretical models have been proposed, they have thus far been difficult to integrate and develop.
In this study, we were inspired in part by grid cells in creating a more general and robust grid module and constructing an interactive and self-reinforcing cognitive system together with Bayesian reasoning, an approach called space-division and exploration-exploitation with grid-feedback (Grid-SD2E). 
Here, a grid module can be used as an interaction medium between the outside world and a system, as well as a self-reinforcement medium within the system. The space-division and exploration-exploitation (SD2E) receives the 0/1 signals of a grid through its space-division (SD) module. 
The system described in this paper is also a theoretical model derived from experiments conducted by other researchers and our experience on neural decoding. Herein, we analyse the rationality of the system based on the existing theories in both neuroscience and cognitive science, and attempt to propose special and general rules to explain the different interactions between people and between people and the external world.
What's more, based on this framework, the smallest computing unit is extracted, which is analogous to a single neuron in the brain. 
\end{abstract}


\section{Introduction}
\label{introduction}
In 1948, Wiener compared the control and communication functions in the machine and nervous system, and established the core position of information and feedback in the cybernetics \cite{Wiener2020cyber}.
In 1950, Alan Turing asked the question, "Can machines think?" \cite{turing1950computing}.
Preliminary answers might be found across various disciplines in neuroscience, cognitive science, the relentless pursuit of technology, and other areas. 
In 1958, Von Neumann came to the important conclusion that the nervous system should be based on two types of communication that involve logical formulas and arithmetical formulas. Here, the former belongs to instruction communication (logical form), and the latter belongs to digital communication (arithmetic form) \cite{Neumann2016the}.
Currently, neural encoding and decoding play an important role in the technical realisation of brain-computer interfaces and in exploring brain mechanisms, mainly for deciphering the relationship between neural brain activity and the external world \cite{wallisch2014matlab, livezey2021deep}. 
In addition, some researchers have also made some important explorations in cognitive science.
Clark proposed the theory of "predictive processing" to perfectly complement the study of embodied minds, and stated that the brain appears to be an action-oriented machine that interacts with the external world to achieve "prediction error minimisation" \cite{Clark2016surfing}.
Friston proposed the "free-energy principle" to explain that brain constantly builds models of the external world, and to better predict changes in the external environment, as well as resist increases in entropy and decreases in cognitive errors \cite{Friston2010active, Ramstead2018Answering}.
Based on the research and observation of the neocortex, Hawkins proposed the theory of "memory-prediction" and "thousand-brain intelligence" to reveal the operation mechanism of the brain \cite{Hawkins2004on, Hawkins2019a, Hawkins2021a}.
Hofstadter was inspired by Gödel's proof \cite{Nagel2001Godel} and proposed the "strange loop" theory and "analogue" idea, trying to explore the deep secrets and cognitive intelligence of
the brain \cite{Hofstadter1999godel, Hofstadter2007i, Hofstadter2013surfaces}.
In recent years, rapid developments and remarkable achievements have been made in both theory and technical application, such as deep learning \cite{Goodfellow2016deep,bengio2021deep}, reinforcement learning \cite{Sutton2013reinforcement,Sutton2022the}.

\begin{figure}   
    \centering
    \includegraphics[width=\textwidth]{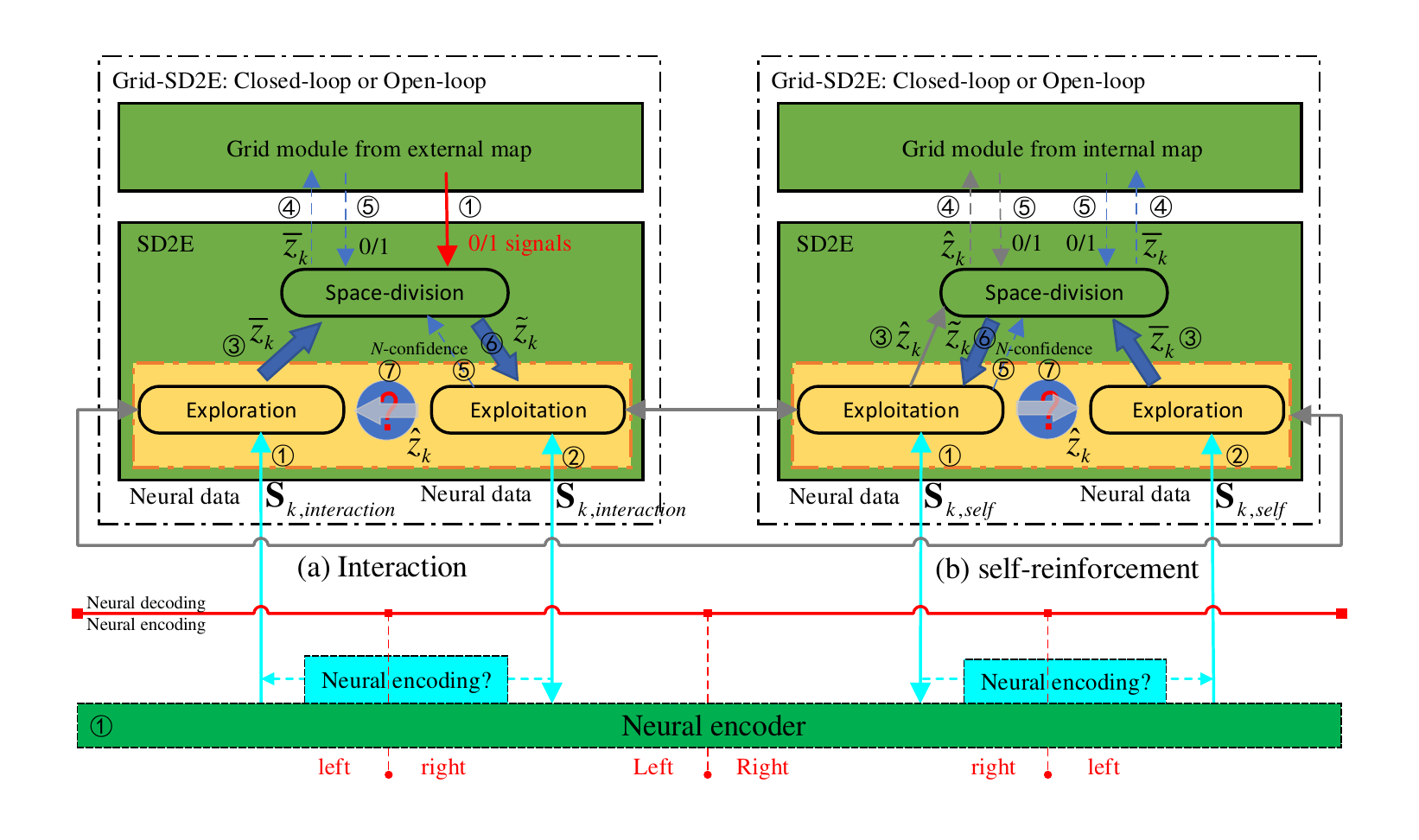}   
    \vspace{-10pt}
    \caption{(Neural decoding) The (a) interaction and (b) self-reinforcement in a cognitive system. 
    If an arrow exists at a red question mark, it is a closed-loop; if not, it is an open-loop.
    In our perspective, the numbering given is only one of many forms that can be seen as a conscious thought and does not represent the entire thinking of the brain. 
    If we consider embodiment (multimodal) and both the conscious and the unconscious, the numbering may differ.
    $\textbf S_{k,interaction}$ and $\textbf S_{k,sef}$ are neural data generated in (a) interaction and (b) self-reinforcement respectively. $z_k$ is the true label. $\bar{z}_k$ is the value decoded by unsupervised exploration. $\tilde{z}_k$ is the value corrected by SD module. $\hat{z}_k$ is the value output by supervised exploitation and is used to update the weights of exploration. $k$ is the $k$th moment or the $k$th sample. 
    The bits (0/1) are the converted symbols after being decoded by SD2E (neural data $\rightarrow$ value decoded $\rightarrow$ 0/1).
    }
    \label{fig: Grid-SD2E_Procedure}
\end{figure}

In addition, we will illustrate the origin of Figure \ref{fig: Grid-SD2E_Procedure}, one of which is a theoretical model derived from experiments and experience on neural decoding. Our purpose is to give it a better theoretical explanation in this paper.
First, we know that the decoded neural data have a time series characteristic \cite{wallisch2014matlab, wu2019neural}. 
In 2018, Feng et al. first discovered the symmetry of the activity space between the decoded unsupervised trajectory and true movement trajectory \cite{feng2018neural}.
In 2020, they conducted a full experimental verification of the symmetry phenomenon, that is, the training trajectory that was corrected once was put back into the training model, and the trained model then reached the level of supervision during the test \cite{feng2020weakly}. 
In 2021, Feng et al. carried out an abstract modelling based on these studies. As their idea, the main model needs to continuously interact with the outside world through an encoded 0/1 signal to correct its own predicted value. Interestingly, this model satisfies the elements of an embodied model \cite{feng2021vif}. 
Finally, we can observe that their embodied model is also able to develop a self-reinforcing capability, as shown in Figure \ref{fig: Grid-SD2E_Procedure}. 
It is worth noting that, because it only has substantial computations of neural data, we tentatively refer to SD2E as the brain in this paper.

Bayesian inference first requires setting some prior state of belief, and then stating how to change that belief based on new information or new evidence; in addition, once one updates one's prior belief to a posterior belief, the posterior belief can also be used as a prior belief at the next observation \cite{Clark2016surfing}. 
As shown in Figure \ref{fig: Grid-SD2E_Procedure} (a), the brain's judgmental predictions (0/1) are compared with perceived signals or observations from the outside world. 
If the results of the comparison are consistent, we will achieve a higher degree of confidence in our cognition (exploitation); 
however, if the results of the comparison are inconsistent, we can revise the cognitive model (exploitation) according to the results of the observations, or believe that the results of the observation are incorrect and then generate behaviour to change the world.
The main goal is to reconcile observations from the external world with the cognitive model of the brain's internal world. 
Here, observations can correct the credibility of the cognitive prior of the background knowledge (exploitation), which forms the background knowledge of a cognitive posterior.
That is, the brain needs to constantly go through a series of interactions, predictions, prior and posterior, among others, to form our lifetime of knowledge or memories (exploitation).
Finally, in Figure \ref{fig: Grid-SD2E_Procedure} (b), in order to apply the background knowledge (from eyes, ears, nose, tongue and body) that has a mapping with the external world, the exploitation replaces the external world to perform the self-reinforcement of Grid-SD2E.

In summary, we imagine that the brain constructs internal world models of how it's sensory inputs are generated. We consider that these models are used to generate predictions of sensory inputs, such that the ensuing prediction errors can be used to update the model when necessary. The particular aspect of these models we explore is that they generate predictions from a discretised or quantal representation of the external world. In other words, we consider discrete state spaces that are suitable for supporting distributed hash codes (see chapter \ref{Distributed hash encoding}). By applying the principles of Bayesian belief updating to these models \cite{Friston2017the}, they can be revised on the basis of prediction errors or used to predict the most likely responses or behaviours. This is known as active inference or predictive processing \cite{Friston2010active, Friston2017active}.

In this context, exploration refers to the optimisation of a model’s structure or parameters – through a process of Bayesian belief updating or learning. In other words, learning is considered a process of exploring model structures and parameters. When the model has been optimised, it can then be exploited to predict the most likely behaviour or actions, in the sense of active (planning as) inference \cite{Attias2003planning, Botvinick2012planning, Da2020active, Lanillos2021active}. One key issue – we focus on below – is the exploration of the model structure. For example, experience-dependent plasticity or neurodevelopment may explore different course graining operators that the determine the number of bins or subspaces in the generative model, whose occupancy is encoded by binary numbers (e.g., the firing of a place cell – or any neuronal population with a well-defined receptive field).

The main contributions of this paper are summarized as follows.
\begin{itemize}
\item We have constructed a cognitive system called Grid-SD2E, which features a general grid-feedback mechanism inspired by a ViF module \cite{feng2021vif} and grid cells \cite{Hafting2005Microstructure}. Additionally, we have proposed a general learning principle that includes both special and general rules. The system can use Bayesian inference to facilitate interaction and self-reinforcement.
\item Unlike Rat-SLAM \cite{milford2004ratslam} and NeuroSLAM \cite{yu2019NeuroSLAM}, which are spatial navigation systems inspired by the path integration of grid cells, the grid module of our proposed system is a coupled analogy with grid cells. This means that it is not only inspired by grid cells but also attempts to derive the specific mechanism behind their formation.
\item We derive and explain the functionality of the grid module based on grid cells. Additionally, we thoroughly analyze the function of SD2E from both theoretical and mental perspectives. 
Ultimately, guided by general learning principles, we aim to extend the system to a range of cognitive tasks. 
\end{itemize}

The rest of this paper is organized as follows.
Section 2 clearly articulates the similarities between grid modules and grid cells, as well as the encoding mechanism and formulation derivation of the grid modules. The operation mechanism and principle of Grid-SD2E are then given comprehensively.
Section 3 deeply explores the ideas of the interaction and self-reinforcement of Grid-SD2E, then we proposed special and general rules.
Section 4 discusses the broader impacts of Grid-SD2E on the original contribution, embodied intelligence, and additional challenges.
Finally, section 5 briefly summarises the main contributions of this study.


\section{The space-division and exploration-exploitation with grid-feedback (Grid-SD2E)}
\subsection{The similarities between the grid module and the grid cell}

\begin{figure*}
\centering
\includegraphics[width=0.98\textwidth]{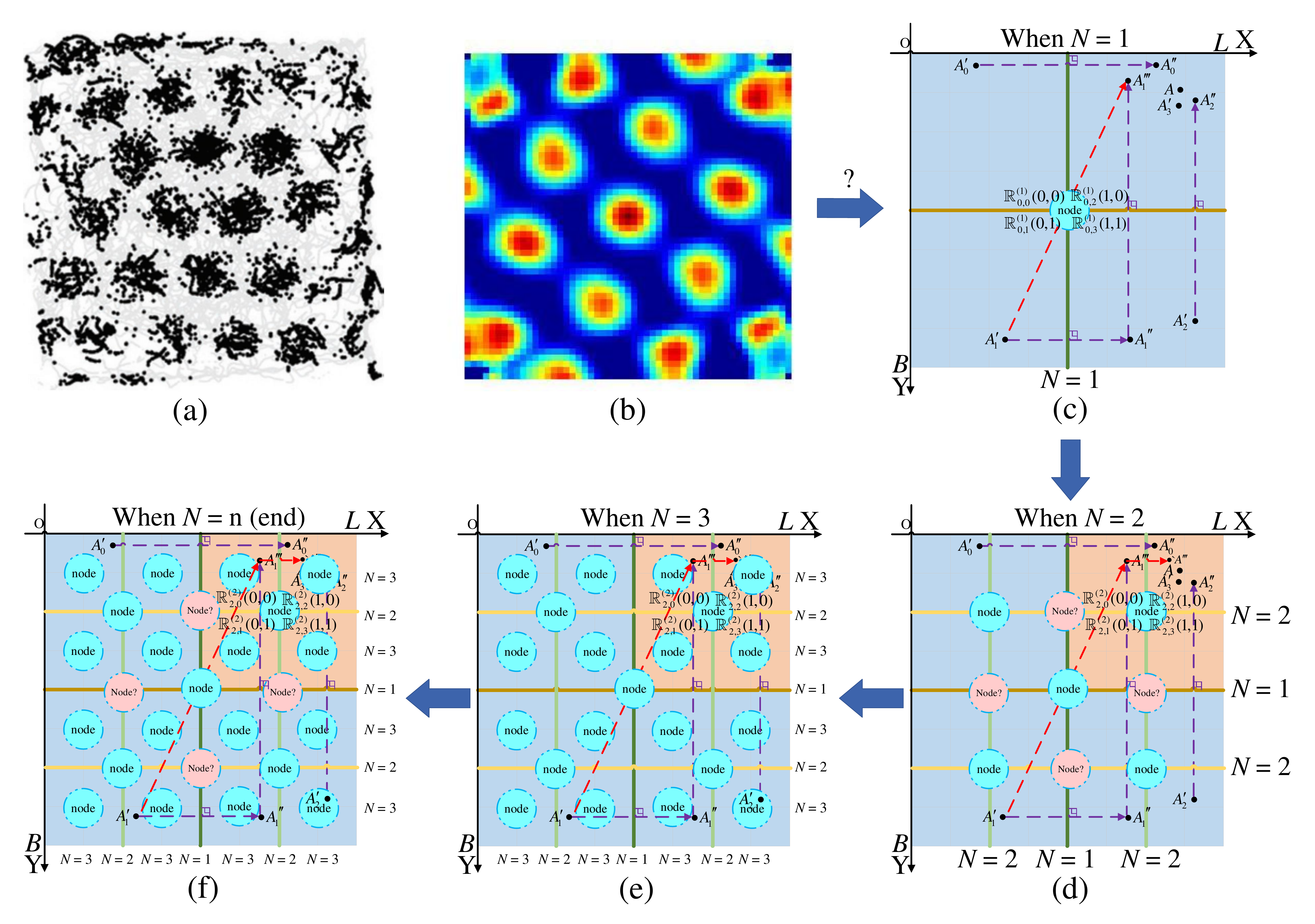}
\caption{Here, picture (a) comes from \cite{moser2014grid}, picture (b) comes from \url{https://commons.wikimedia.org/wiki/File:Autocorrelation_image.jpg}, and picture (c) comes from \cite{feng2021vif}.
(a) Grey shows the movement trajectory of a foraging rat in a 2.2m wide square enclosure. 
Black shows the spatial firing pattern of a grid cell from the rat medial entorhinal cortex. 
Each black dot is one spike of grid cell when the rat moves to that location.
(b) When rat moved in a certain space, a more precise result was obtained after statistical denoising of the firing positions of a grid cell. Regions with warmer colors (i.e. red) were more densely activated, while regions with cooler colors remained repressed. 
(c) Under the condition of parameter $N=1$, the 0/1 signal encodes the position within the moving space and correct the predicted position when applied to this space. Node is the intersection of the virtual $x-$ and $y-$axes when $N=1$.
(d), (e) and (f) The encoding and correction of the predicted position in each sub-space is similar to that in (c). In the same way, each sub-space generates a new sub-node. The pink node with a question mark represents a filling node.
}
\label{fig: Grid-SD2E_movementspace}
\end{figure*}

In 2005, Hafting et al. found that when rat foraged within the range of activities, the entorhinal cortex of the brain had "grid cell" similar to map positioning, and the grid cell was more ordered \cite{Hafting2005Microstructure}.
Note that in 1971, O'Keefe et al. found that when rat ran to a certain area, the hippocampus of the brain had "place cell" that always showed an activated state \cite{keefe1971the}; subsequently, "Head direction cell" \cite{taube1990head} and "Border cell" \cite{Lever2009boundary} have also been discovered one after another.
The grid cell needs to be anchored to a landmark as a reference, and when the landmark is rotated, the grid structure of the grid cell will also rotate \cite{Hafting2005Microstructure}. 
Moreover, when the external space becomes larger, the grid of the grid cell is expanded, and the spacing becomes larger; in contrast, when the external space becomes smaller, the grid of the grid cell is compressed, and the spacing becomes smaller \cite{Barry2007experience,Stensola2012the}.
The space boundary generates a shear force on the grid of a grid cell, causing the grid axis to deflect at a certain angle. For example, when the rat first enters an unfamiliar space, the grid axis is perpendicular to the boundary; deflection will not occur until the rat is completely familiar with the space \cite{Stensola2015shearing}.
The rat was placed into two connected compartments. Initially, the rat generated a separate "grid map" for each room. When the rat was thoroughly acquainted with the two spaces, the two separate grid structures merged into one, which means the change from local to whole \cite{Giocomo2015spatial}.
The space encoded by two grid cells has some grid-scale, grid orientation, and grid phase characteristics \cite{moser2014grid}.
Researchers have found that the structure presented by grid cell in 3-D space is not as orderly and regular as that illustrated in 2-D space. Still, the local distance between the firing positions of each grid cell is a substantial order, and regular. They suggested that grid cells may play a more integral role in the formation, processing, and consolidation of memory \cite{Ginosar2021local, Grieves2021irregular}.

The cellular structures distributed in the hippocampus and entorhinal cortex may jointly construct one or multiple coordinate systems and allow us to understand complex cognitive navigation and memory functions better.
In Figure \ref{fig: Grid-SD2E_movementspace} (a) and (b), these clusters appear to be distributed on a straight line and have symmetry about the center of activity space. 
In Figure \ref{fig: Grid-SD2E_movementspace} (c), (d), (e) and (f), if $N$-value is larger, the resolution of a certain position in the space is also larger; if $N$-value is smaller, the positioning resolution is smaller.
The similarities between the grid module and grid cells are:
(1) Both have their activity spaces and boundaries and satisfy symmetry or approximate symmetry for the center of activity space, respectively.
(2) In the grid modules with parameter $N>1$ and the multi-cluster grid cell, both have their multiple intersections, respectively. When the $N$-value is changed in the grid module and the size of the space is changed in the grid cell, the spacing of their respective intersections also changes respectively.
(3) The local and global characteristics were mentioned in the 2-D and 3-D grid cell studies, and the grid module is also a whole or global space that integrates multiple local subspaces at $N>1$.
Finally, based on the above similarities and observations, we empirically derived the general formula of the grid module. 

\subsection{The encoding mechanism and formula derivation in the Grid/SD module}

\begin{figure*}
\centering
\includegraphics[width=0.90\textwidth]{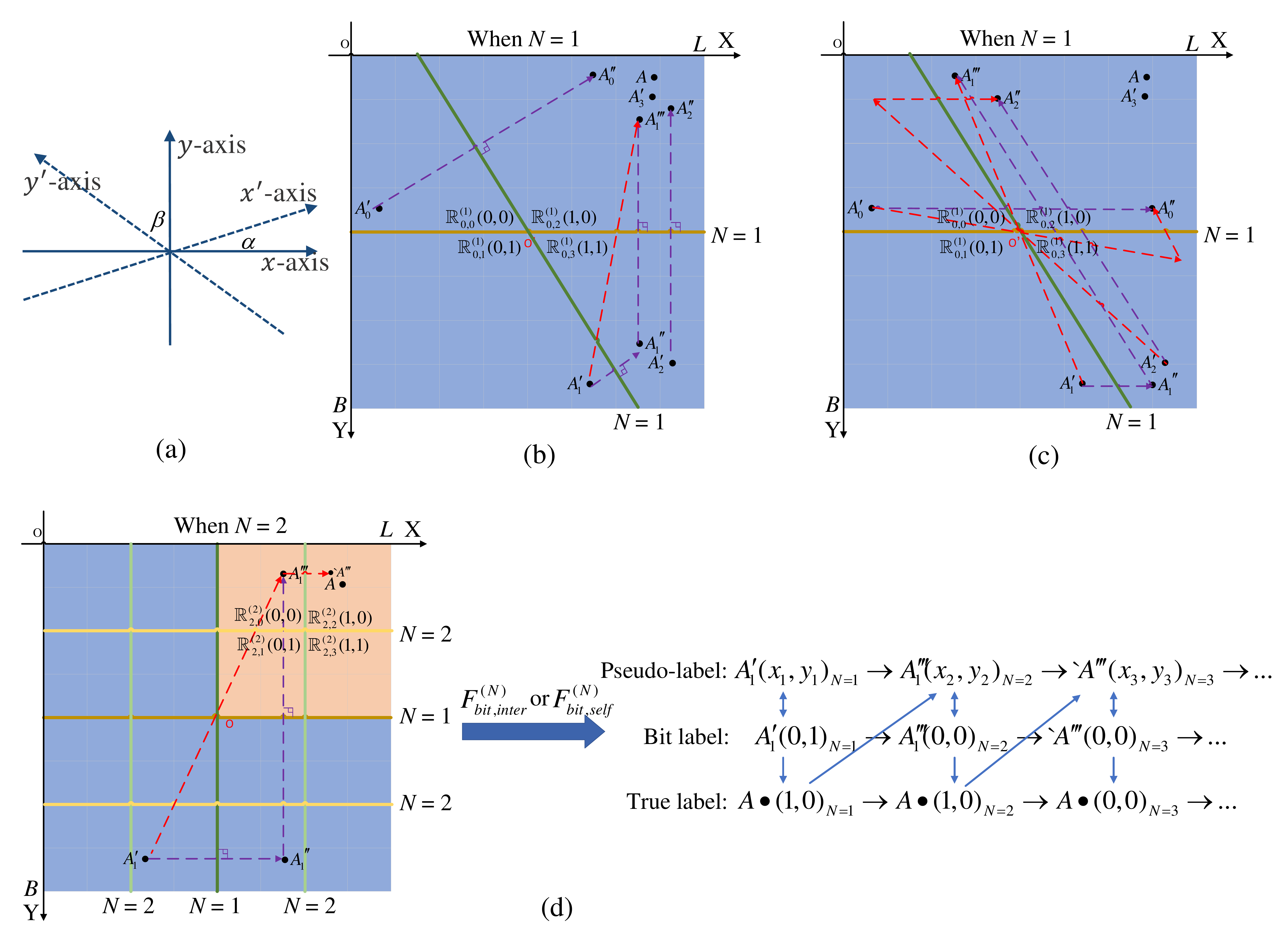}
\caption{(a) Suppose the deflection angle of the $x$-axis is $\alpha$ and for the $y$-axis is $\beta$ (mimicking the tilt shown by the grid cell map in Figure \ref{fig: Grid-SD2E_movementspace}b).
(b) and (c) are two schemes given under the condition (a), whose purpose is to keep the predicted values and the true labels in the same subspace. Here, the red line is only used as an auxiliary line for reference, and the blue area is an active space.
(b) When $x$- and $y$-axes are deflected to become the $x'$- and $y'$-axes, the lines between the true position and the position of correction are still perpendicular to the $x'$- or $y'$-axes, in appearance.
In addition, given the true label $A\bullet$-coordinate, the possible predicted values $A'_0\bullet$, $A'_1\bullet$, $A'_2\bullet$ or $A'_3\bullet$ are about to be symmetrically flipped based on being perpendicular to the deflected $x'$- and $y'$-axes, and the $\bullet A''_0$, $\bullet A'''_1$, $\bullet A''_2$ and $\bullet A'_3$ are the values obtained after being symmetrically flipped, respectively.
(c) When $x$- and $y$-axes are deflected to become the $x'$- and $y'$-axes, the lines between the true position and the position of correction parallel the $x'$- or $y'$-axes.
It's worth noting that all values are equal on either line parallel to the $x'$- or $y'$-axes.
(d) Finally, the diagram illustrates the step of correcting pseudo-labels by drawing an analogy between the bit (0/1) labels generated from unsupervised predictions and the bit (0/1) labels  of the true position.
}
\label{fig: Grid-SD2E_movementspace-2}
\end{figure*}

\subsubsection{Distributed hash encoding in the grid module}
\label{Distributed hash encoding}

Distributed hash encoding (or spatial encoding) mainly includes both distributed and hash encoding.
Distributed encoding means that data $A\bullet$-coordinates can be represented by combinations of 0s and 1s in an active space.
The advantage is that the representation vector is low-dimensional and dense, different from high-dimensional sparse one-hot encoding (indicating that the vector must include a 0 or 1 to refer to certain data).
Hash encoding refers to the assumption that data $A\bullet$-coordinates (or binary-coded data) of any length are input, and after being processed by a hash function, a fixed-length hash value $key$ (combination of 0s and 1s) is output, that is, $Hash(A\bullet)= key$.
This process is irreversible; that is, $A\bullet$coordinate cannot be reversed by the $key$.
Specifically, refer to Figure \ref{fig: Grid-SD2E_movementspace-2} and the movement encoder \cite{feng2021vif}.
Here, $\mathbb R$ represents the movement space.
When $N=0$, this is itself such as $\mathbb R$: $\mathbb R^{(0)}(0/1, 0/1)$.
When $N>0$, subspace $\mathbb R^{(N)}_{n, i}(x_{bit}, y_{bit})$ is the $i$th quadrant of the $n$th subspace of the $N$th divided space, where $N$ denotes the $N$th division of the space, $n$ is the $n$th subspace, $i$ is the $i$th quadrant, and $x_{bit}$ and $y_{bit}$ are the encoded $x$- and $y$-positions in the $xy$-plane.

Finally, the inspiration provided by the grid module in this paper is that in a complete space, any position can be encoded as a 0/1 string in a distributed form; the other is that this encoding mode also satisfies the characteristics of hash encoding.

\subsubsection{Formula derivation for SD module interacting with grid module}
\label{Formula derivation}

Figure \ref{fig: Grid-SD2E_movementspace-2} show two alternatives, (b) and (c), under condition (a), where the $x$- and $y$-axes are deflected. Figure \ref{fig: Grid-SD2E_movementspace-2} (d) is that he diagram illustrates the step of correcting pseudo-labels by drawing an analogy between the bit labels generated from unsupervised predictions and the bit labels of the true position.
First, if we pursue simplicity, efficiency, and low power consumption in the brain, we believe that the theoretical derivation of (b) is more complicated than (c) and is less likely to form the grid nodes analyzed, similar to grid cells, in the Figure \ref{fig: Grid-SD2E_movementspace}. The formula derivation of (b) is presented in 'Appendix A'.
The main text presents a derivation of (c). These formulae are mainly based on the following aspects:
(1) The deflection angles of the $x$- and $y$-axes have changed, and the origin of the self-built coordinate system is the center of the motion space, then $\eta_1 = \eta_2 =1$ in 'Appendix A'.
(2) The deflection angles of the $x$- and $y$-axes have changed, and the origin of the self-built coordinate system is any position of the motion space, then $\eta_1$ and $\eta_2$ in 'Appendix A'.
(3) The deflection angles of the $x$- and $y$-axes have changed, the origin of the self-built coordinate system is any position of the motion space, and all values are equal on either line parallel to the $x'$- or $y'$-axes.
Among them, these factors have a progressive relationship in our theoretical analysis and derivation, and the formula in 'Appendix A' satisfies factors (1) and (2).
Finally, it's worth noting that all values are equal on either line parallel to the $x'$- or $y'$-axes in the self-built coordinate system of factor (3). Additionally, under low complexity conditions of factor (3), grid nodes that are more easily generated form a coupling analogy with grid cells. This also appears to provide insight into the formation mechanism of grid cells described in this paper.

Therefore, we empirically provide the formula for factor (3) in space $\mathbb R$. 
That is, the calculation formulas for the movement encoder and corrector \cite{feng2021vif} are rewritten as follows: 
\begin{equation}
    {\lim\limits_{{\mathbb R}: N\rightarrow +\infty}} F^{(N)}_{bit, inter}=\left\{
             \begin{array}{llr}
                1\; ; \quad if \; N=j, \; and, \; \overline z_k \;or\; z_k \geq f_{mid}(\bullet) \\
                0\; ;  \quad if \; N=j, \; and, \; \overline z_k \;or\; z_k  < f_{mid}(\bullet)
             \end{array}
    \right.
\end{equation}
\begin{equation}
    {\lim\limits_{{\mathbb R}: N\rightarrow +\infty}} F^{(N)}_{bit, self}=\left\{
             \begin{array}{llr}
                1\; ;  & if \;  N=j, \; and, \; \overline z_k \;or\; z_k > f_{mid}(\bullet) + \epsilon \;\\
                0\; ;  & if \;  N=j, \; and, \; \overline z_k \;or\; z_k < f_{mid}(\bullet) - \epsilon \\
                dropout \; or \; F^{(N)}_{bit, inter}\; ; & others \\
             \end{array}
    \right.
\end{equation}
\begin{equation}
    {\lim\limits_{{\mathbb R}: N\rightarrow +\infty}} F^{(N)}_{update}=\left\{
             \begin{array}{llr}
                \overline z_{k}\; ;  & if \; N=j, \; and, \; \overline z_{k,bit}=z_{k,bit}  \\
                F^{(N)}_{correct}\; ;  & if \; N=j, \; and, \; \overline z_{k,bit} \neq z_{k,bit}
             \end{array}
    \right.
\end{equation}
where, formula (1) is for encoding the external world during interactions. 
${\lim_{{\mathbb R}: N\rightarrow +\infty}} F^{(N)}_{bit, inter}$ indicates that, as $N$-value increases in space $\mathbb R$, any position encoded can approach the subspace where the true label is located.
Among them, $F^{(N)}_{bit, inter}$ is the encoded 0/1 strings. For example, the predicted $\bar z_{k}$ and true label $z_{k}$ are respectively encoded as $\bar z_{k, bit}$ with 0/1 strings and $z_{k, bit}$ with 0/1 strings.
In addition, $j=0, 1, 2, ..., N$, and $f_{mid}(\bullet)$ is the central axis (or reference line) of the spatial symmetry. 
If $\overline z_k \;or\; z_k \geq f_{mid}(\bullet)$, the movement is encoded as 1; otherwise, the movement is encoded as 0. 

Formula (2) is for encoding the internal world during self-reinforcements.
${\lim_{{\mathbb R}: N\rightarrow +\infty}} F^{(N)}_{bit, self}$ indicates that, as $N$-value increases in space $\mathbb R$, any position encoded can also approach the subspace where the true label is located.
Among them, $F^{(N)}_{bit, self}$ is the encoded 0/1 strings as $\bar z_{k, bit}$ and $z_{k, bit}$. 
If $\overline z_k \;or\; z_k > f_{mid}(\bullet) + \epsilon$, the movement is encoded as 1; if $\overline z_k \;or\; z_k < f_{mid}(\bullet) + \epsilon$, the movement is encoded as 0; otherwise, the movement is dropped out or enters the interaction $F^{(N)}_{bit, inter}$.
$\epsilon$ is cognitive or uncertainty bias. That is, the probability of incorrect predictions at the edge of the divided boundary line (or reference line) is greater or uncertain in self-reinforcement due to factors such as noise interference, so $dropout$ and $F^{(N)}_{bit, inter}$ needs to be introduced, similar to forgetting (or human forgetfulness) and avoiding forgetting (or human interaction with the outside world to enhance memory).

\begin{figure}    
    \centering
    \subfigure[Local method in Grid-SD2E]{
        \includegraphics[width=0.98\textwidth]{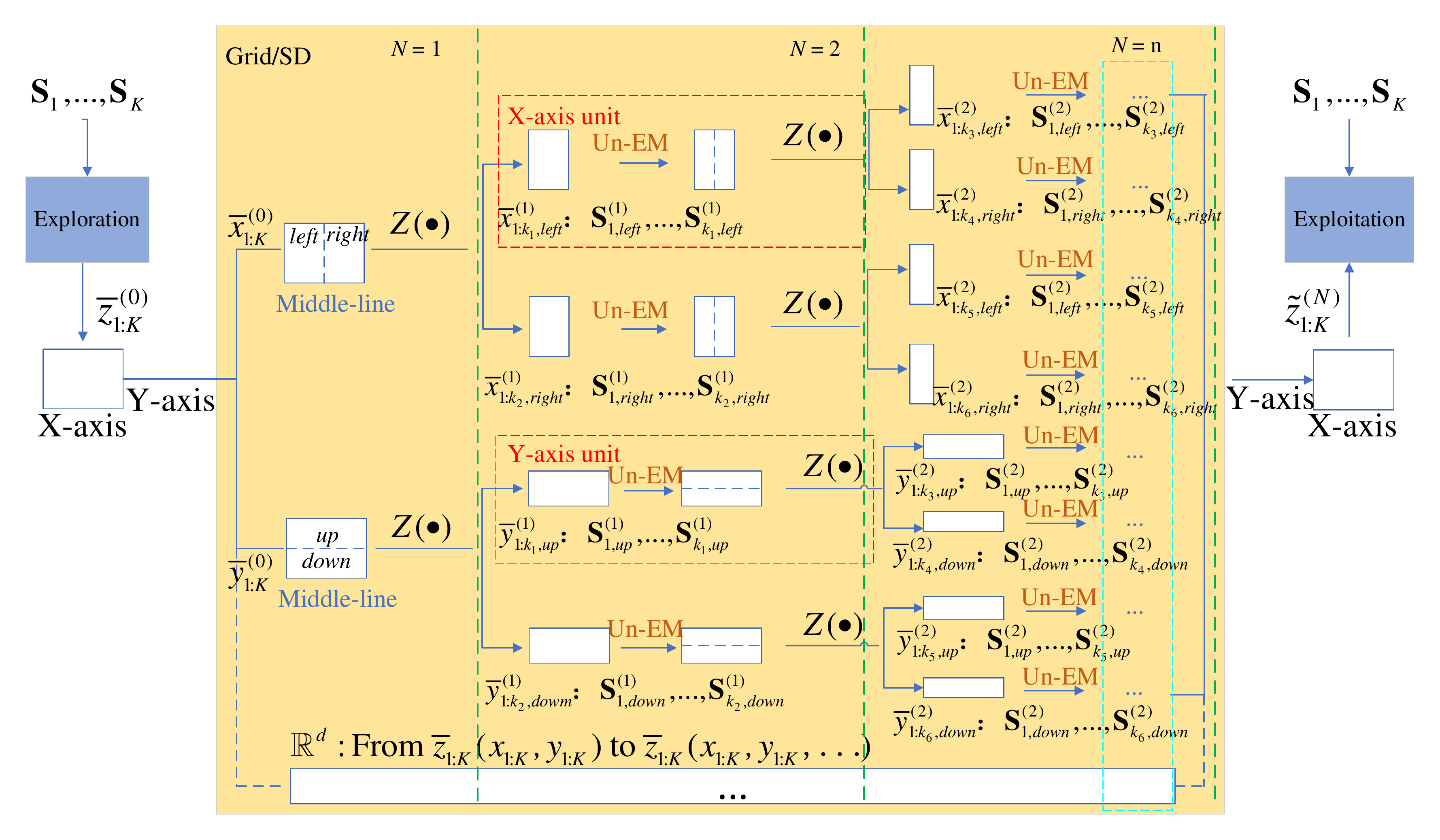}
    }
    \subfigure[Local unit in Grid-SD2E]{
        \includegraphics[width=0.86\textwidth]{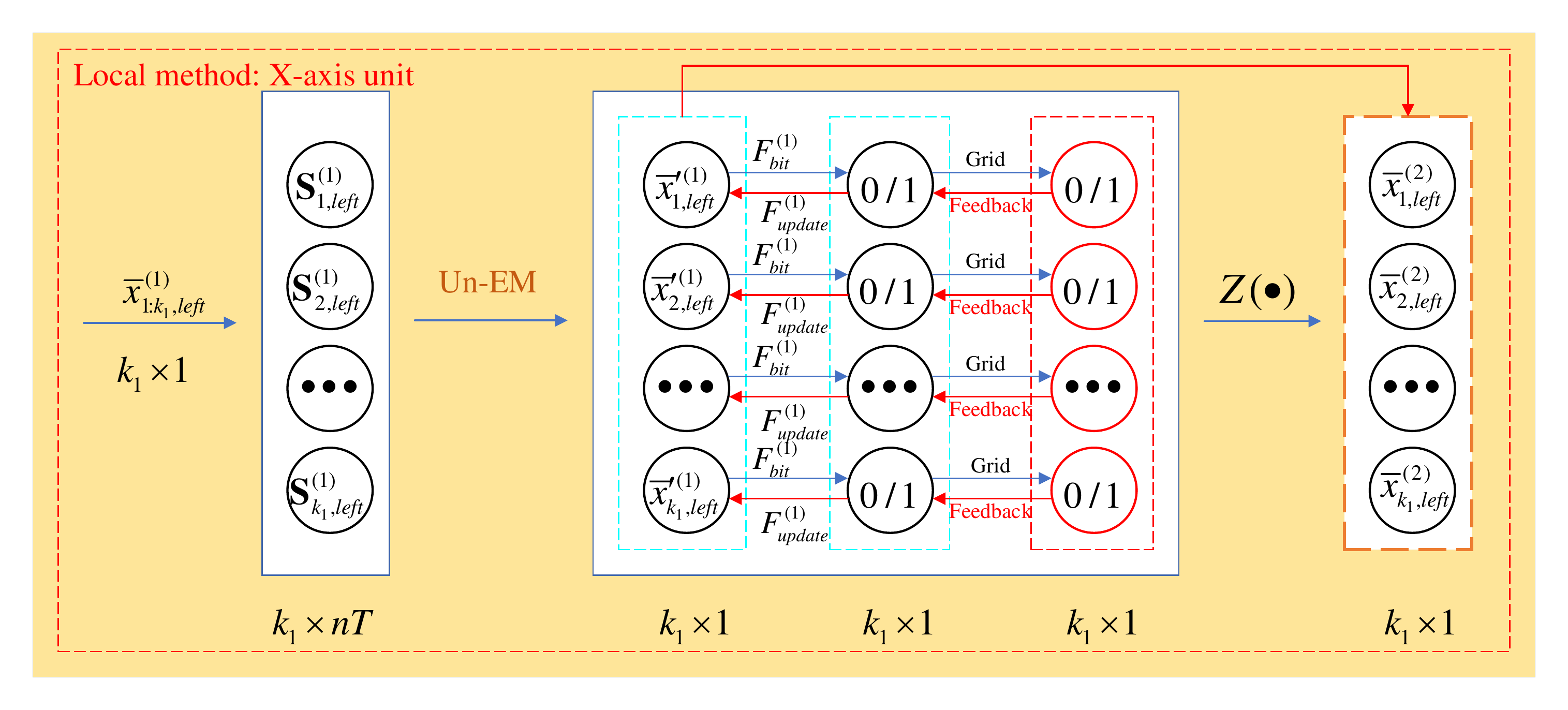}
    }
    \label{fig: Grid-SD2E_local}
    \caption{(a) The steps of the local method used in Grid-SD2E \cite{feng2021vif}. The red box represents the smallest processing unit, called a local unit in a cognitive system.
    The local method is concerned about whether the neural data $\textbf S^{(1)}_{1:k_1, left}$ corresponding to $\overline x^{(1)}_{1:k_1, left}$ is reused in the local unit to obtain $\overline x'^{(1)}_{1:k_1, left}$, instead of the source $\overline x^{(1)}_{1:k_1, left}$.
    Furthermore, the inputs of the local method are $\overline x^{(0)}_{1:K}$ and $\overline y^{(0)}_{1:K}$ predicted by the unsupervised exploration, which are then also applied in Eqs. (1) and (2) for encoding and correction.
    The output of the local method is the corrected $\tilde x^{N}_{1:K}$ and $\tilde y^{N}_{1:K}$, which, along with the corresponding neural data, are also sent to the supervised exploitation for training.
    (b) The processing unit of the local method \cite{feng2021vif}.
    Here, $\overline x'^{(1)}_{1:k_1, left}$ instead of $\overline x^{(1)}_{1:k_1, left}$ are encoded as 0/1 signals in the subspace, and are then compared with the given external 0/1 signals to obtain the corrected $\overline x^{(2)}_{1:k_1, left}$. 
    }
    \label{fig: Grid-SD2E_diagram}
\end{figure}

Formula (3) is for correcting incorrect predictions during interactions and self-reinforcements.
${\lim_{{\mathbb R}: N\rightarrow +\infty}} F^{(N)}_{update}$ indicates that, as the $N$-value increases in space $\mathbb R$, any position corrected can approach the true label.
Among them, $F^{(N)}_{update}$ and $F^{(N)}_{correct}$ are updated (or corrected) values. 
Beside, for ease of representation in Figure \ref{fig: Grid-SD2E_diagram}, we define the function $Z(\bullet)$ to represent the role of $F^{(N)}_{bit, inter}$ and $F^{(N)}_{update}$ (or $F^{(N)}_{bit, self}$ and $F^{(N)}_{update}$), without any additional implications.

Next, under the condition of factor (3), the calculation formula for $F^{(N)}_{correct}$ is given as follows.
In addition, because the reference centers on the $x'$- and $y'$-axes of a space have different calculation modes, respectively, the predicted values on the $x$- and $y$-axes have different correction formulas.
\begin{equation}
    F^{(N)}_{correct}=\left\{
             \begin{array}{lr}
                \overline x_{k} + 2\lambda_1 * f_x(\eta_1, x_{\max}, x_{\min}, \overline x_{k})\; ;  &  if \; N=j, \; \overline z_{k} = \overline x_{k} \\
                \overline y_{k} + 2\lambda_2 * f_y(\eta_2, y_{\max}, y_{\min}, \overline y_{k})\; ;  &  if \; N=j, \; \overline z_{k} = \overline y_{k} 
             \end{array}
    \right.
\end{equation}
where, when $\overline z_{k} = \overline x_{k}$, $f_{mid}(\bullet) = \frac{(x_{max} + x_{min})}{2}\eta_1$; when $\overline z_{k} = \overline y_{k}$, $f_{mid}(\bullet) = \frac{(y_{max} + y_{min})}{2}\eta_2$. 
$z_{max}$ and $z_{min}$ are the maximum and minimum boundaries of movement during the $N$th division, respectively.
$\lambda_1$ and $\lambda_2$ are the offset rates of the symmetric distances along the $x'$- and $y'$-axes, respectively.
$\eta_1$ and $\eta_2$ are the offset rates of the reference center along the $x'$- and $y'$-axes, respectively.
$f_x(\eta_1, x_{\max}, x_{\min}, \overline x_{k}) = \frac{(x_{\max} + x_{\min})}{2}\eta_1 - \overline x_{k}$ is the deviation of the predicted $\overline x_{k}$ from the tilted centre on the $x'$-axis of the moving space.
Similarly, $f_y(\eta_2, y_{\max}, y_{\min}, \overline y_{k}) = (\frac{y_{\max} + y_{\min})}{2}\eta_2 - \overline y_{k}$ is the deviation of the predicted $\overline y_{k}$ from the tilted center on the $y'$-axis of the moving space.

Finally, when $N \rightarrow +\infty$, Grid-SD2E is equivalent to a supervised mode, and when $N=0$, it is an unsupervised mode. 
It should be noted that in Eqs. (4), the deflection angles ${\alpha}$ and ${\beta}$ are eliminated under the condition of Figure \ref{fig: Grid-SD2E_movementspace-2} (c). 
For example, the angle between the $x'$-axis and the $y'$-axis can be any value (e.g., 60). 
In addition, it is worth noting that when we only consider the reference center as a symmetry, the predictions on the $x$- and $y$-axes are directly independent in Figure \ref{fig: Grid-SD2E_movementspace-2} (c) and Eqs. (4), respectively. 
Based on our analysis, the similarity between grid modules and grid cells is demonstrated in that corrective (or firing) behavior is more likely to occur at nodes in theory. This is because incorrect predictions of location are more likely to happen at the node (reference center) on the edge of the reference line due to factors such as noise interference. 
If this derivation is reasonable, then it is possible that the brain has developed a self-built coordinate system (including corrective firing, etc.) that is functionally equivalent to the Cartesian system but in a different form that better aligns with the external world.

\subsection{Spatiotemporal mining of the Grid-SD2E on cognitive system}
\label{Grid-SD2E-overview}

\begin{figure*}
\centering
\includegraphics[width=\textwidth]{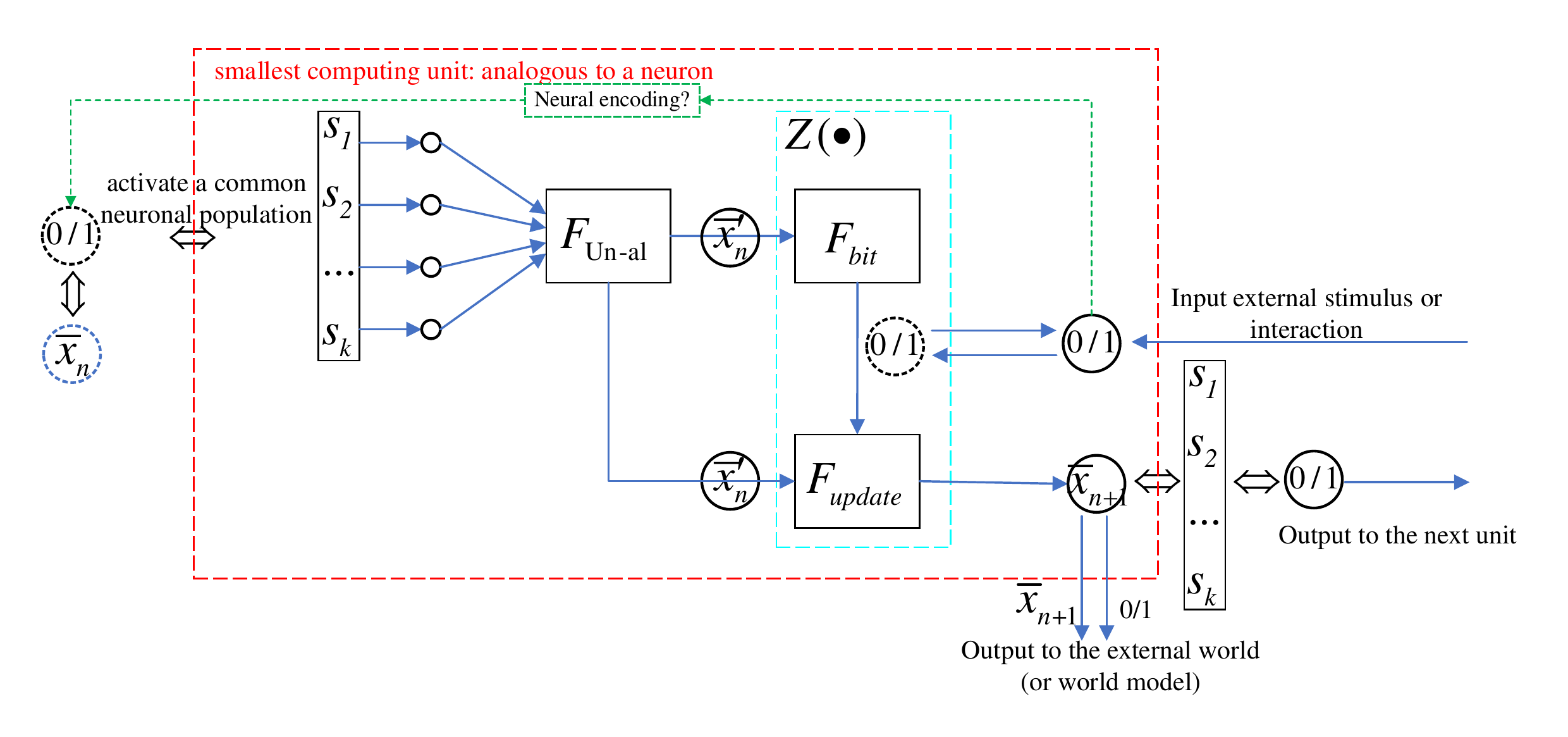}
\caption{The Smallest computing unit is from Figure \ref{fig: Grid-SD2E_diagram}, which can be compared to a single neuron.
}
\label{fig: Smallest_computing_unit}
\end{figure*}

\subsubsection{An introduction to the function of the Grid-SD2E}
\label{An introduction}

As the main feature of Grid-SD2E in Figure \ref{fig: Grid-SD2E_Procedure}, the Grid/SD module utilises robust 0/1 signals within the activity space/subspace to judge and process the values predicted through an unsupervised exploration, and then sends them to a supervised exploration for training. 
In Bayesian inference, exploitation with a embodied memory is used to evaluate the adaptive $N$-confidence and to carry out the interaction and the self-reinforcement.
Please refer to Figures \ref{fig: Grid-SD2E_Procedure} and \ref{fig: Grid-SD2E_diagram} for details of the system. 
Here, Figure \ref{fig: Grid-SD2E_diagram} (a) shows the local approach of Grid-SD2E in a space and subspace. 
The term local indicates that the decoded values of the neural data reused in each subspace replace the predicted values entering this subspace, rather than processing only the predicted value entering the space. 
Refer specifically to the local unit in Figure \ref{fig: Grid-SD2E_diagram} (b). Grid-SD2E is composed of more extended local units when applying a local method. Here, $k_1 \times nT$ has $k_1$ rows and $nT$ columns, where $k_1$ is the number of samples in this subspace, $n$ is the number of features of each sample, and $T$ is the previous $T$ moments in the current state of each sample.
Finally, when $N = 0$, Grid-SD2E is an unsupervised exploration without an interaction. 
When $N = 1$, we consider the predicted values to be satisfactory if the encoded 0/1 signals in Grid-SD2E match the given external 0/1; if not, we flip the predicted values symmetrically to keep them both in the same subspace (see Figure \ref{fig: Grid-SD2E_diagram}). 
When $N = 2$, the operation is similar to that of $N = 1$, and so on. 
In our view, as infants grow and face new experiences, they need to increase their $N$-value to increase their self-attention or supervision; however, when they have learned knowledge and reasoning, they are more inclined toward the unsupervised level under prior experience, i.e., $N$ is small. For details, please refer to \cite{feng2021vif}.

In addition, as shown in Figures \ref{fig: Grid-SD2E_movementspace} and \ref{fig: Grid-SD2E_diagram}, as the $N$-value increases, the resolution of the positioning continually increases. 
However, the number of subspaces increases exponentially, resulting in an exponential increase in the number of computations. 
Based on the power consumption of the brain, the $N$-value should be as small as possible, indicating that the prediction of the internal system is more closely matched with the external world. 
Here, if $N$ is larger, it indicates that the neural data generated by the brain are more chaotic and cannot be effectively used to make predictions, that is, they are disordered. 
Of course, if $N$ is smaller, the neural data generated by the brain are more accurate and orderly. 
Therefore, we believe that when the brain minimises $N$ in Grid-SD2E, the neural data generated by the brain should be transitioning from disorder to order.
Next, we analyse "how a group of neurons forms a behavior". 
When the brain interacts with the external world, neural data are generated by a group of neurons, which are then decoded into predicted values and encoded as 0/1 signals. 
A 0/1 signal encoded from the brain is analogised with the encoded 0/1 signal from the outside world; here, the analogy results in a population of neural data being assigned to different subspaces. 
Finally, the self-organised neural data are separately predicted, encoded, and analogised in their respective subspaces from $N = 1$ to $N = n$. 
From the above, in Grid-SD2E, we believe that a group of neural data go through the self-organisation of the brain to complete the output of the behavior.
In addition, from the perspective of Grid-SD2E, it is simple and convenient to expand from a 2-D space to a $d$-D space in the self-organising system (assuming that the prediction in each dimension is independent in the conclusion of Figure \ref{fig: Grid-SD2E_movementspace-2}c.), see Figure \ref{fig: Grid-SD2E_diagram}. 
\begin{equation}
    \mathbb R \xrightarrow[]{+d} \mathbb R^d
\end{equation}
where, the space $\mathbb R$ can be denoted as $\mathbb R^d$, and $d$ represents the dimension of the space. 
In our view, compared to the 2D biological navigation system using grid cells, the advantage of extending to multi-dimensional biological systems is that on the one hand, these seem to be beneficial for our navigation activities in the three-dimensional world, and on the other hand, for cognitive tasks other than spatial navigation, multi-dimensional systems may be more conducive to their execution of distributed parallel computing. 
In addition, since each dimension is assumed to be independent, it further reduces the complexity of the system to perform spatial navigation and other cognitive tasks.

\subsubsection{Self-organizing properties of the Grid-SD2E on cognitive system}

Finally, to draw an analogy with a single neuron in the brain, the smallest computing unit is showcased in Figure \ref{fig: Smallest_computing_unit}, from Figure \ref{fig: Grid-SD2E_diagram}.
This is a computing unit, which can do self-correction function by inputting 0/1 feedback signals.
$[s_1, s_2,..., s_n]$ are the neural firing signals corresponding to the location $\overline x_n$, and $F_{\rm Un-al}$ refers to any unsupervised algorithm (Un-al).
$\overline x'_n$ is the predicted position obtained by the $F_{\rm Un-al}$ on the neural data $[s_1, s_2,..., s_n]$.
$\overline x'_{n+1}$ is the corrected predicted position.
The computing unit has two outputs: 
the first is the neural data (or 0/1) corresponding to the corrected prediction value, which is then sent to the next computing unit; and the second is the corrected prediction value or 0/1 symbols, which are used to determine prediction errors based on the exploitation or be outputted directly in the last layer.

From Figure \ref{fig: Grid-SD2E_diagram} and Figure \ref{fig: Smallest_computing_unit}, it can be concluded that the self-organization characteristics of neural data in grid-SD2E are mainly manifested in the following case: when moving from $N=n$ to $N=n+1$, the neural data will follow the decoding position of a certain subspace, while adjusting the prediction target. 
The decoding position is corrected via an analogy between the encoded 0/1 signals and the external feedback 0/1 signals.
Of course, it is uncertain whether the neural data corresponding to the decoding position would undergo re-neural encoding or change in each subspace. 
It seems that validation is required from computational neuroscience and neurophysiology.

\subsubsection{Correlation between the Grid-SD2E and the free energy principle}
\label{free energy principle}

Furthermore, if we focus on the number of subspaces or bins (increasing exponentially with $N$) that support a distributed hash encoding, there are some fundamental principles that suggest there will be an optimum value for $N$. These principles underwrite the self-reinforcement or self-evidencing inherent in the free energy principle \cite{Hohwy2014the}. Self-evidencing corresponds to self-reinforcement by maximising the ‘marginal likelihood’ of predicted sensory inputs. This marginal likelihood is also known as ‘model evidence’ and can always be expressed as accuracy minus complexity. The complexity term is important, because it means that self-evidencing (i.e., self-reinforcing exploitation) requires an accurate explanation that is as minimally complex or parsimonious as possible.

This means, for any interaction or world there will be an $N$ (at each level of a hierarchical model) that maintains accuracy, which is as small as possible. This tendency to optimally coarse-grain or tile state spaces can be seen from several perspectives. From the perspective of the free energy principle, it inherits from Jaynes maximum entropy principle \cite{Jaynes1957information}; in the sense that free energy approximates marginal likelihood and comprises an entropy term (which has to be maximised) and some constraints (supplied by the generative model). An alternative rearrangement of entropy and constraints is in terms of accuracy and complexity, as above. Here, complexity corresponds to the KL-divergence between priors and posteriors. In other words, complexity scores the degrees of freedom used to provide an accurate account of sensory data. It is this that we associate with the number of subspaces that should be kept as small as possible to minimise complexity cost. This is also known as Occam's principle and is closely related to formulations of universal computation in terms of Solomonov induction and Kolmogorov complexity \cite{Schmidhuber1991curious, Schmidhuber2010formal, Sun2011planning}. This leads to alternative formulations of variational free energy in terms of minimum description and message length \cite{MacKay1995free, Wallace1999minimum}. These are interesting perspectives because again, they speak to minimising $N$ under constraints. In short, the current formalism we are proposing speaks in a natural way to information theoretic treatments of optimal learning and inference.
Furthermore, the Grid-SD2E system still necessitates a similar theoretical treatment formula as follows. 
\begin{equation}
    {\lim\limits_{{\mathbb R^d}: N\rightarrow +\infty}} F^{(N)}_{update} \xrightarrow[]{minimising \; N} {\lim\limits_{{\mathbb R^d}: N\rightarrow 0}} F^{(N)}_{update}
\end{equation}

where, the operation of minimizing $N$ is equivalent to reducing the length of the hash encoding at a specific position in space, in order to decrease complexity and enhance the accuracy of prediction.

In 1944, Schrödinger proposed that "life exists with negative entropy" concept \cite{Schr1944what}.
Friston's free-energy principle is based on Bayesian reasoning (the prior becomes a posterior, which becomes a prior again); in addition to the "free-energy minimisation (i.e. the difference between the expected state of the internal world and the measured state of the external world)" while interacting with the environment, the free-energy principle also introduces an "information theory isomorphism of the thermodynamic free-energy" to explain the existence of life as negative entropy \cite{Friston2010active, Ramstead2018Answering}. 
If Grid-SD2E is correct regarding the information processing, we believe that it is in line with Schrödinger's and Friston's perspective. 
That is, the neural data generated from disorder to order are resistant to an "entropy increase" in making optimal predictions and reducing cognitive errors. 
Then, when neural data are transmitted in a system, they can be self-organised into group signals of different subspaces for prediction.

\section{The in-depth exploration of the interaction and self-reinforcement of cognitive system}

\subsection{Some key work and inspiration in cognitive science}

In 1931, Gödel proposed that "incompleteness theorem" \cite{Godel1931uber}. 
In Gödel's proof \cite{Nagel2001Godel}, a consistent formal system must be incomplete and needs to be judged as "true" from outside the system, which constitutes the core of self-referentiality.
Next, we focus on Hofstadter's strange-loop theory\cite{Hofstadter1999godel, Hofstadter2007i}.
The strange-loop theory can be understood in terms of the framework presented in this paper: it is a combination of rules, and has more than one layer of the high-level to be used to judge the low-level; the occurrence of intelligence is equivalent to judging whether the high-level can be generated autonomously from the formal system (low-level). 
In the strange-loop, "I" is an epiphenomenon of an epiphenomenon, and "I" is the best way to interact with the external world under natural selection; the reason why "I" is me lies in the "dancing" pattern of symbols in the brain. 
In addition, the "loop" is the feedback loop. 
The "strange" has two points: one is the active thinking (some complex cognitive processes will occur when the brain receive information, such as screening, analogy, etc.); the second is the ignorance of the underlying mechanism (the brain contains a self-referential feedback loop---actively thinking about external information via rich symbols, and in turn constantly expanding the set of symbols.).

In Pearl's causality theory\cite{Pearl2009causality, Pearl2018the}, Pearl divides the cognitive ability into three levels using the "ladder of causality." 
The first layer is a correlation, whereby observers establish statistical correlations in their own consciousness. 
They subsequently make predictions by observing, analyzing data, and identifying patterns. 
The second layer is intervention, where the observer combines observation with active change, discards the passive acceptance of data, adjusts the causal probability map in consciousness to establish the causal relationship, and then takes remedial measures. 
The third layer is counterfactual reasoning, whereby observers are required to construct an imaginary world, reflect, and abduct upon the past. 
Events, and infer the ways in which the relevant results will change in this case to verify whether the causal relationship is established.

\subsection{The special rule and general rule of the Grid-SD2E on cognitive system}
\label{special and general}

\begin{figure*}
\centering
\includegraphics[width=\textwidth]{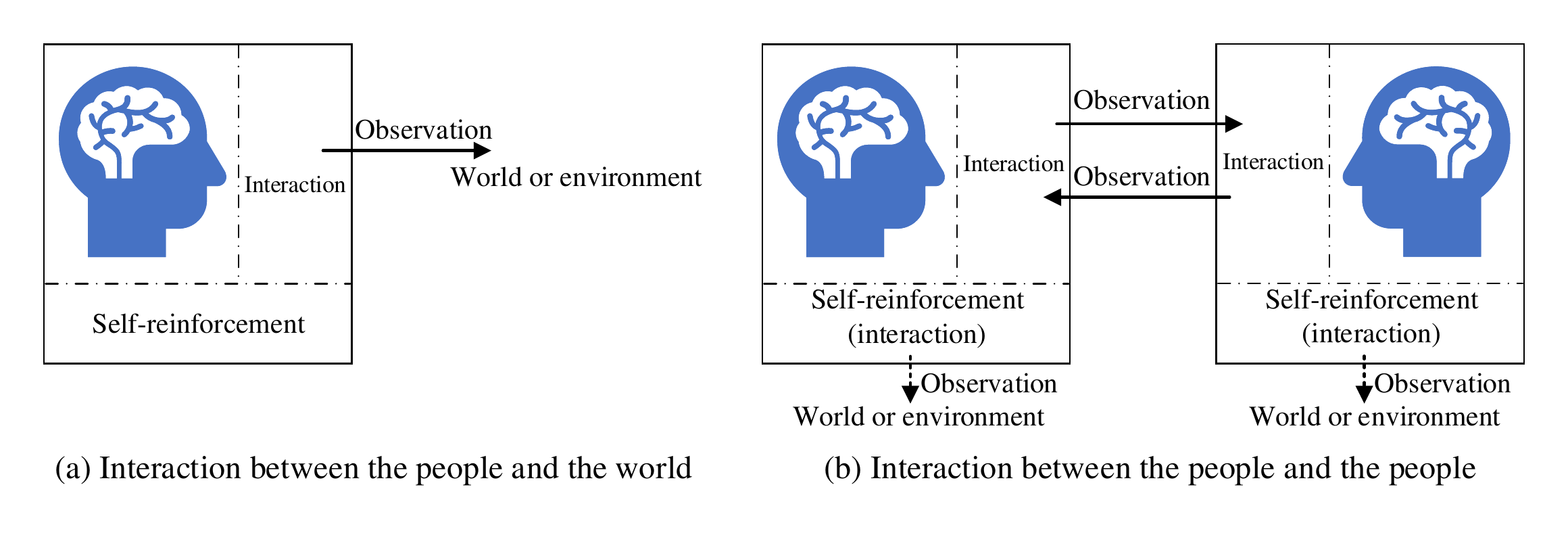}
\caption{(a) In the special rule, the subject (people) interacts with the object (world or environment), excluding interactions with other subjects. 
(b) In the general rule, the subject interacts with the other subjects, and the interaction between the subject and the object will degenerate into the self-reinforcing ability of the subject.
}
\label{fig: Grid-SD2E_movementspace-4}
\end{figure*}

\emph{Special rule:} The cognitive reinforcement = interaction (subject and object) + self-reinforcement (subject); e.g., I read the book first to form a memory (interaction), then I recite it without looking at the book (self-reinforcement), but I forget something while reciting, and then reread the book (interaction), next I recite it (self-reinforcement), etc. 
\emph{General rule:} The cognitive reinforcement = interaction (subject and other subjects) + self-reinforcement (subject, or subject and object); e.g., I communicate with others and we don't work well together (interaction), I think about ways to communicate (self-reinforcement), I look for knowledge from the outside world and rethink the ways (self-reinforcement = object + subject), then I communicate with them and we understand each other very well (interaction), etc. Reference subject (people), and object (real-world or environment) in Figure \ref{fig: Grid-SD2E_movementspace-4}.

The difference between the special and the general lies in whether the circulation of information is one-way or two-way.
The level of two-way circulation is the highest (subject + other subjects), and the level of one-way circulation is medium (subject + object), while the level without circulation is the lowest (subject or object). 
In the special rule, the subject and the object are at different levels, and the subject acquires knowledge from the object, while in the general rule, the subject and other objects are also at different levels. Despite that the transmission of information is mutual, it is not simultaneous, and either party can acquire knowledge from the other party (depending on the scene, time and place, etc.). 
At this time, the subject and the object degenerate into the same level, and the object is equivalent to the body of the subject. However, they still conform to special rule. 
Furthermore, in the special rule, the communication between interaction and self-reinforcement is realized through knowledge or memory acquired from the object, whereas in the general rule, the communication between interaction and self-reinforcement is realized via their respective values, beliefs and knowledge, etc., obtained from the object and other subjects. 
It is worth noting that in the general rule, 
other subjects share the object with the subject, but they never belong entirely to the subject. 
In other words, the subject has two bodies, one of which is the perception and the other is the external world (object), but only the perception is unique to the subject.

If Grid-SD2E is correct regarding the information processing, it can be inferred that the "0/1" string encoded in Grid-SD2E may refer to the "I" of the strange-loop, for the reason that "0/1" is an epiphenomenon of an epiphenomenon of neural data (neural data $\rightarrow$ value predicted $\rightarrow$ 0/1). 
We also believe that the two-way transmission of information is the foundation and reason why one are intelligent.
Other subjects can relieve the confusion of 0/1 generated by the subject, and can also give an explanation for 0 or 1 (or $\rho_1$\textbf{0}, $\rho_2$\textbf{1}). Moreover, vice versa. In this case, $\rho_1$ and $\rho_2$ are probabilities of 0 and 1 respectively, for example, people use probabilistic words like "must, should, can, may and others" in our language and beliefs.
Regardless of the special or general rules, self-reinforcement (low-level) cannot solve or confirm the problem, and interaction (high-level) is required to provide solution to it.
In addition, self-reinforcement is also a strange-loop in itself and the refined knowledge/experience (high-level) from the outside world and other subjects are adopted to guide the other information (low-level) to make judgments.
Last but not least, based on the above analysis, in order to overcome the incompleteness of one's own knowledge, one need to continuously learn knowledge from the outside world (object) and other subjects.
In other words, if there is no communication, there is no intelligence, that is, communication promotes intelligence under certain conditions. 
Here, we give the rules of communication about the mind, which we believe are also key to developing intelligent models.
\begin{equation}
    'I' (0/1, \rho_1\textbf{0}, \rho_2\textbf{1}) \autorightleftharpoons{?}{} {'others'} (0/1, \rho_3\textbf{0}, \rho_4\textbf{1})
\end{equation}
where, $\rho_1, \rho_2$ and $\rho_3, \rho_4$ are the probabilities of 0 and 1 respectively in 'I (self)' and 'others (world and people)', and their values are from 0\% to 100\% in probability or strength of belief. 
In addition, a detailed interpretation and analysis (they were added after the comments had been completed.) of the $\rho_1, \rho_2$, $\rho_3$, and $\rho_4$ can be found in section \ref{paradoxes}.
We believe that this formula is closely related to the development of the mind and the alignment of value.
Therefore,  we cite the three principles of human-machine function and ability allocation \cite{Liu2021integrated}, referred to as "Three Laws of Human-Machine" on the basis of friendly interaction:
\begin{itemize}
\item A robot should not disturb humans when humans are more confident;
a robot should alert humans when humans are overconfident;
a robot should help humans when humans are not confident.
\end{itemize}
on the other hand, we believe that if a robot also possesses the characteristic of being unsure, which requires collective consciousness (or the joint consciousness of all robots) to jointly determine, it may be able to get along better with humans. In other words, the limitations that humans have should also be selectively possessed by robots. 

In addition, the first layer in causality can be answered with facts through data analysis, and the latter two layers must use causal models. 
In Grid-SD2E, given the evidence from the outside world, the "exploration $\rightarrow$ SD $\rightarrow$ exploitation" in interaction passively accepts data, thereby satisfying the first layer. 
When the "exploitation experienced $\rightarrow$ SD" in interaction is directly invoked to intervene (it can also be said to yield the $N$-value satisfying the brain.) and take action, it satisfies the second layer. 
When we invoke the exploitation reflected in self-reinforcement for simulated reasoning, it satisfies the third layer.
As inferred from Figure \ref{fig: Grid-SD2E_Procedure}, the exploitation experienced in interaction and the exploitation reflected in self-reinforcement are identical in form but sometimes differ in function. 
Here, what is different from Pearl's causal ladder is that it can simulate intervention (Here, simulated intervention can be distinguished from experiential intervention at the second layer.) and counterfactual reasoning in the third layer, that is, action can subsequently be taken after a comprehensive analysis of the pros and cons in self-reinforcement.
Judging from the above three layers, if one only has experience and do not reflect or imagine, then one is not different from the external world, because experience directly comes from the external world in this paper.
Here, this reflection is a reflection of pursuit of rationality.
In other words, the reflection on the third layer may be one of the main features that are different from non-life substances.

Therefore, strange-loop seems to be Gödel's or Hofstadter's feedback loop from the inside out at the cognitive level, whereas causal theory is Pearl's correlation, intervention and reflected reasoning from the outside in at the cognitive level. 
Finally, an incomplete system in Gödel's 'incompleteness theorem' needs to continuously expand to a larger system, until there is a superfinite system that can cover all problems and approach the truth. 
Similarly, in the special and general rules, if a brain is regarded as an information processing system
, then the two-way circulation can avoids the one-way superfinite expansion.

\subsection{Some of our thinks on the Grid-SD2E and autonomous system}

First, in our opinion, a 'world model' that stores background knowledge and common sense is necessary \cite{Ha2018world, Friston2021world, Hawkins2021a, A2022LeCun}. 
In Grid-SD2E, the exploitation is equivalent to a 'world model' or a 'mental model' that store, update, and recall information from the 'eyes, ears, nose, tongue, and body.'
In addition, from the perspective of our systems, some constructive perspectives are given as below on intelligence.
\subsubsection{Turing Test}
On the basis of 'Turing Test', if the abilities are divided into skills and intelligence \cite{wang2007introduction}, the skill-value can be judged through information communication and observation, which is also called the term "Embodied Turing Testing \cite{Zador2023catalyzing}". However, the limit of intelligence depends on the internal design of the system, including hardware (framework: Grid-SD2E) and software (within the framework), which determines the degree of intelligence-value that can be achieved. In other words, intelligence depends on the efficiency of information dissemination when multiple tasks are performed in a designed learning system, and it is then reflected through skills \cite{Chollet2019on, Hawkins2021a}. 
Furthermore, whether this design is ‘immortal computation’ or ‘mortal computation’ \cite{hinton2022the}, and whether the knowledge base that stores knowledge is transferable, still needs further discussion.
The following literature can be taken as reference: General intelligence's information processing principles, New benchmarks for intelligence levels, and The memory-bearing cerebral cortex that implies intelligence \cite{wang2007introduction, Chollet2019on, Hawkins2021a}; Four element of cognition (matter, energy, structure, and time) about the relationship between hardware and software \cite{li2023cognitive}.
\subsubsection{Subjectivity and objectivity}
Intelligent communication should contain both subjective and objective aspects. Subjectivity is one's own feelings in emotion, such as “I believe”, “I think”, “I feel”, etc., which plays a role in evaluating, screening and accepting the factual data. Objectivity is something that is learned directly in rationality, and the factual data can be acquired through manual programming and online search. Objectivity does not involve subjective consciousness. The literature below is for reference: 
The 'calculation' of rational logic is learned in form through Western thinking, and the 'plan' of perceptual logic is learned in terms of change through Eastern thinking, also referred to as 'jisuanji' theory \cite{Liu2021integrated}; the determination of whether an AI system possesses consciousness is primarily based on the similarity between the computational functionalism of the AI system and that proposed in a scientific consciousness theory (Recurrent Processing Theory, Global Workspace Theory, Higher-Order Theories, Attention Schema Theory, Predictive Processing, Agency and Embodiment, etc.) \cite{Butlin2023consciousness}; in addition, there are theories incompatible with computational functionalism, such as integrated information theory \cite{Oizumi2014from, Tononi2015Consciousness, Albantakis2021what, Michel2020on, Mediano2022the}; a low entropy state of self-consciousness, i.e. 'Boltzmann Brain' \cite{Bostrom2002anthropic}.  
\subsubsection{Value and belief}
What are value and belief? Beliefs may initially arise from surprise and unusual information, while value may be drawn from later input factual data that is for or against such surprise or unusual information. The former is belief that can be set artificially or acquired autonomously, and the latter is value acquired by the system itself. For example, the "Three Laws of Robotics" and other derived laws are considered to be set and cannot be violated in beliefs ($\rho_1$ and $\rho_2$ can represent the strength of belief.). Regarding this aspect, please refer to the following literature: Categorization by analogy, The theory of thousand-brain intelligence in the cerebral cortex, and Ideas for building core frameworks that can embed human knowledge \cite{Hofstadter2013surfaces, Hawkins2021a, Marcus2021insights}; what is trustworthy machine intelligence (MI)? Some researchers believe that it should have sixteen abilities (Explanation, Deduction, Induction, Analogy, Abductive Reasoning, Theory of Mind, Theory of Mind, Meta-knowledge and meta-reasoning, Quantifier-fluency, Modal-fluency, Pro and Con Arguments, Defeasibility, Explicitly ethical, Sufficient speed, Sufficiently Lingual and Embodied, Broadly and Deeply Knowledgeable.) and their combinations (Planning, Learning.) \cite{Lenat2023getting}. 

\subsection{Paradoxes in Grid-SD2E and logical linguistics}
\label{paradoxes}

Finally, we give some remarks about the paradox on cognitive system of this paper. 
For example, "I am unprovable" in the liar's paradox may be equivalent to "0/1 am 0" in the rational world. 
However, in the emotional world, they may be equivalent to the following.
\begin{equation}
\begin{split}
    'I \; am \; unprovable' \xLongleftrightarrow[]{s.+v.+o.} {'0/1 \; am \; \rho_1\textbf{0}'} \; (or\; {0/1 \; am \; \rho_2\textbf{1}}) \\
    \xLongleftrightarrow[]{s.+v.+o.} {'0/1 \; <must, should \; and \; others> \; \textbf{0}'}
\end{split}
\end{equation}
As can be seen, "0/1 am 0" is a special case of "0/1 am $\rho_1\textbf{0}$", that is, $\rho_1=100\%$ in the emotional world. Therefore, we should be in the emotional world, not the rational world, most of the time.
In other words, rationality is relative to emotion, and rationality in mathematical or symbolic logic may be a special case embedded in emotion. 
Based on the above analysis, it can also be a cognitive game with embedded rationality and emotion in Eqs. (7)–(8).

Furthermore, to identify rationality and emotion more clearly, in this paper, we believe that rationality can be explained as the "knowledge law of Bayesian probability" that follows the subjective prior \cite{Hoang2018La}, and emotion can be defined as the "spiritual law of conscious experience" that follows ethics and morality. 
Rationality contains subjective (priori probability) and objective (Bayesian probability), and emotion also includes subjective (morality of self-discipline) and objective (ethics of heteronomy).
What's more, this book describes Yin and Yang in the Book of Changes. 
In this paper, some concepts are defined:
Yin is 0, and Yang is 1;
subjective Yin is $\rho_1$ (0\% to 100\%) Yin, and objective Yin is $\rho_3$ (100\%) Yin;
subjective Yang is $\rho_2$ (0\% to 100\%) Yang, and objective Yang is $\rho_4$ (100\%) Yang.

\section{Discussion}
\subsection{The original contribution of the designed Grid-SD2E}
\label{contribution}

Grid-SD2E is the result of our research over many years \cite{feng2018neural, feng2020weakly, feng2021vif}.
Refer to the original collected data set: https://booksite.elsevier.com/9780123838360/ $\rightarrow$ Chapter Materials $\rightarrow$ Chapter 22 \cite{wallisch2014matlab}.
Based on this framework, the smallest computing unit is extracted, which is analogous to a single neuron in the brain. 
In addition, on the theoretical basis of Gödel's 'incompleteness theorem' and Hofstadter's 'strange-loop theory', two interaction modes are proposed ('special rule and general rule'). 
In both the special and general rules that include an interaction and self-reinforcement mechanism, the free-energy principle can reduce predictive errors and then save energy in maintaining the system with low power consumption. The causality 3-layer theory provides the system with cognitive reasoning abilities.
What's more, these also seem to be regarded as a computational theory.
In Marr's "computation theory" \cite{marr2010vision}, the brain is regarded as a complex information processing system involving three layers: the computing layer, the algorithm layer and the implementation layer. 
It is emphasized that 'computational theory' is the key to understanding the brain.

Furthermore, in order to continue to verify the system in theory and experiment in the future, two foundational hypotheses are refined and constitute Grid-SD2E's underlying ideas.
One hypothesis is that the encoded (or generated) neural data meet a symmetrical attribute between the unsupervised decoding trajectory and the true trajectory in the activity space; the other hypothesis is that the cognitive system is an incomplete system that requires non-stop interaction and analogy.
Among them, the first hypothesis is a meta-hypothesis discovered in brain neural data for spatial navigation, while the second hypothesis is developed based on the basic model/framework (ViF-SD2E and Grid-SD2E) of the first hypothesis.

\subsection{Embodied intelligence and controlled hallucinations}
Where does the controlled hallucination come from?
Embodied and distributed cognition have been described in cognitive science \cite{Hatch1993finding, Hutchins1995cognition}.
Predictive coding \cite{Friston2010active, Clark2016surfing} holds that our subjective experience is not a direct reflection of external reality.
Its underlying assumption is that a visual system tries to construct an internal system of the external world and uses that system to actively predict incoming signals.
What one experiences, to a certain extent, is the world expected to be experienced. 
In our observation, high-dimensional information from the external world is compressed and reduced to 0/1 signals by the grid module, which then enter the SD2E for the prediction and inference of the brain. 
Of course, this may also be the reason for the robustness of cognitive system.

In Grid-SD2E, we believe that each 0/1 signal (class) as an analogy object contains a large number of predicted or reference values that can be used as a category or associative library for the build of the system.
Because the characteristics exhibited by the 0/1 signals seem to be more in line with 'categorization by analogy'. 
Here, the 0/1 signal varying from $N=0$ to $N \rightarrow +\infty$ is also equivalent to the transition from fuzzy state to concrete state. 
For example, apples include apples in fruit stores and apples in mobile phone stores.
Based on the analysis of other articles \cite{Marcus2003the, Mitchell2019a}, if 0/1 is regarded as a symbol in this paper, we believe that the 0/1 are likely to be an inherent abstract pattern generated through interaction or feedback in the special rule and general rule.

\subsection{The additional challenges of the Grid-SD2E system}
\label{challenge}

In the specific implementation of the system, we believe that the most important thing is how neural data with this symmetry attribute are generated, that is, how neural encoding in the brain generates such neural data. 
This is the basis for the system to be fully autonomously executed, because in our view, the generated neural data in steps 1 and 2 of Grid-SD2E in Figure \ref{fig: Grid-SD2E_Procedure} could be external perceptual encoding and internal predictive encoding, respectively.
Further, what is the relationship between perceptual encoding and predictive encoding, and how is their relationship realized? In our thinking, the distribution of generated neural data and hidden noise may play a key role. 
In addition, in the spatial domain, the system relies more on the encoding and discrimination of 0/1 for spatial navigation; in the non-spatial domain, the system seems to rely more on 0/1 with probability for various cognitive tasks. Furthermore, we believe that both domain modes should be able to be analyzed and displayed in the geometric grid space. 

\section{Conclusion}
\label{conclusion}

In this paper, we constructed a cognitive system with a general and robust grid-feedback, called Grid-SD2E.
A cognitive system together with Bayesian inference is a theoretical model derived from multiple experiments conducted by other researchers and our experience on neural decoding, including interaction and self-reinforcement. 
In addition, we were partially inspired by grid cells to further derive a more general grid module, the role of which is to mainly guide the external interaction of the system and its internal self-reinforcement. We also analysed the rationality of the system from the existing theories such as neuroscience (free-energy principle and life exists with negative entropy), cognitive science (strange-loop of self-reference, causality 3-layer theory), and so on.
In our opinion, there is a process of resisting 'entropy increase' for the neural data that are generated from disorder to order in cognitive system; and Gödel's 'incompleteness theorem' also seems to imply the incompleteness of cognitive system. Therefore, by applying general learning principles through two kinds of interactions in special and general rules, one can expand their knowledge and understanding.
Finally, if Grid-SD2E is a reasonable description of information processing, we believe that it can be used as a reference for simulating the mind and developing more understanding intelligent systems.
Especially important, prediction, analogy, and self-reflection (or imagination) may be the core of life thinking.

In the future, it still needs to be further explored based on computational neuroscience whether the function of the smallest computing unit of Grid-SD2E is similar to that of real neurons.
For example, self-organizing and neural encoding from disordered to ordered distributions, etc.
Especially noteworthy is that a neural decoding framework (Grid-SD2E) is proposed in this paper, while further exploration is needed for a neural encoding framework to better match the decoding framework.

\section*{Acknowledgements}
The authors would like to express our gratefulness to Karl J. Friston (University College London) for his suggestions and comment on the paper from the perspective of free energy principle, which is of great help to improve the quality of the paper. Finally, we would like to thank some anonymous peers in the field for their invaluable guidance in describing these ideas and expanding further research.

\medskip
{
\small

\bibliographystyle{unsrt}

}

\newpage
\begin{figure}   
    \centering
    \includegraphics[width=0.86\textwidth]{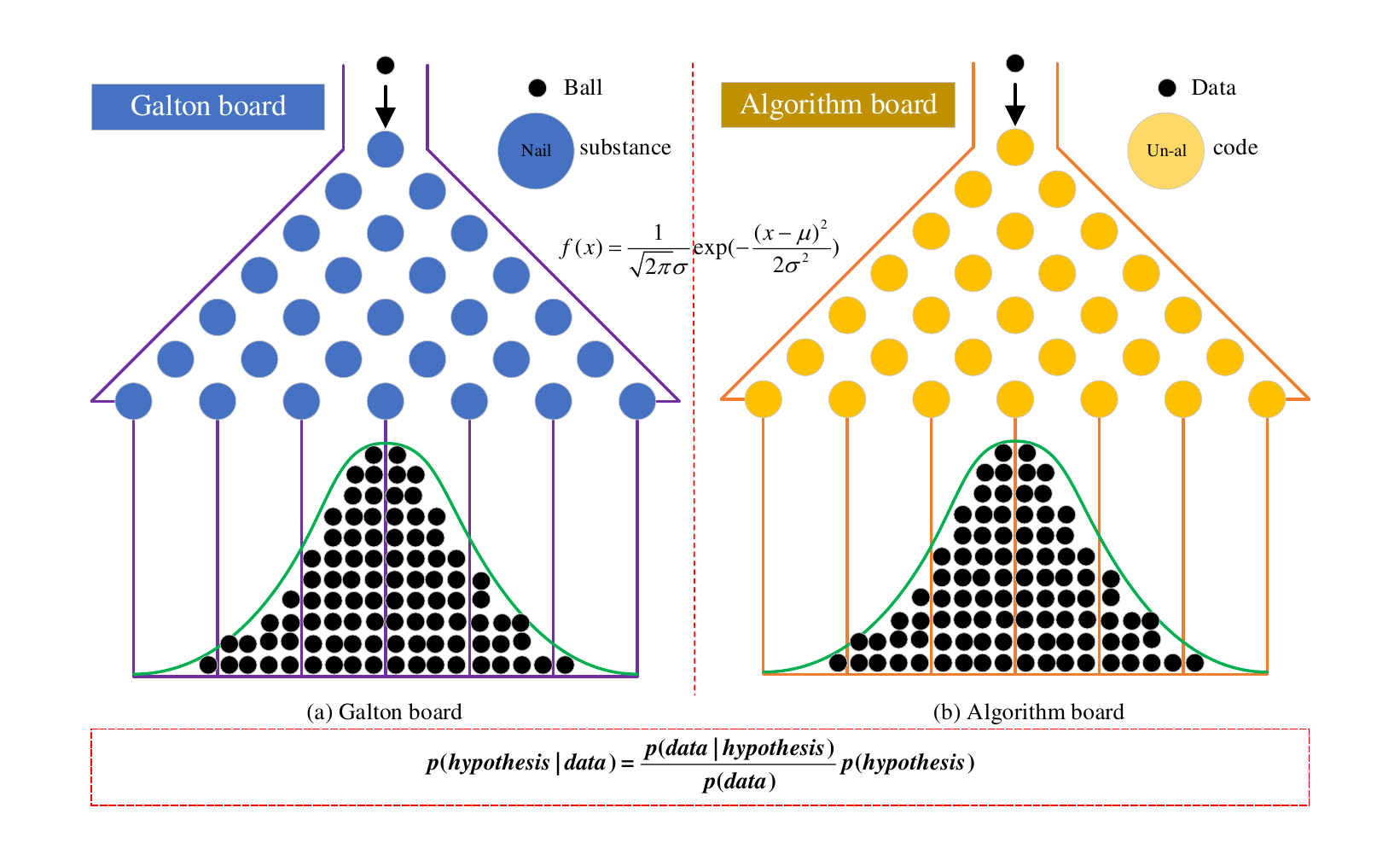}   
    \vspace{-8pt}
    \caption{A Galton board and an algorithm board [5].
    }
    \label{fig: Grid-SD2E_board}
\end{figure}

\textbf{\textit{(1) MUST contain an explanation of the novel contribution the paper makes to the current published literature.}}

Ours: In previous studies, we discovered certain symmetry [1, 2] in neural signals decoded by unsupervised methods in motor, and built a cognitive system [3, 4] based on this pattern. However, the distribution state of data flow that has a significant impact on neural decoding positions remains a puzzle within the system. This further limits the development of the interpretability of the system. See the abstract for a detailed statement [5]. In other words, the validation results of the paper [5] confirm the validity of the previous studies [1, 2, 3, 4]. From the perspective of classical mathematical statistics theory, specifically the Galton Board, it represents the mathematical basis for the discovered symmetry. Consequently, the derived algorithm board constitutes the mathematical underpinning of the system being investigated.

\textbf{\textit{(2) MUST contain details of the prior expertise of the author the paper makes to the current published literature.}}

Ours: In this line of research, all the conclusions reached and those inferred [1, 2, 3, 4] stem from the symmetry discovered in neural decoding. Collectively, the paper [5] essentially constitute the fundamental groundwork for the system; or these works essentially constitute the fundamental groundwork for this research direction. 


\begin{thebibliography}{10}

\bibitem{Wiener2020cyber}
N.~Wiener.
\newblock {\em Cybernetics: or the Control and Communication in the Animal and
  the Machine}.
\newblock Peking University Press, 2020.

\bibitem{turing1950computing}
A.M. Turing.
\newblock Computing machinery and intelligence.
\newblock {\em Mind}, 59(236):433--460, 1950.

\bibitem{Neumann2016the}
J.V. Neumann.
\newblock {\em The Computer and the Brain}.
\newblock Yale University Press, 2012.

\bibitem{wallisch2014matlab}
P.~Wallisch, M.E. Lusignan, M.D. Benayoun, Baker T.I., A.S. Dickey, and N.G.
  Hatsopoulos.
\newblock {\em MATLAB for neuroscientists: an introduction to scientific
  computing in MATLAB}.
\newblock Academic Press, 2014.

\bibitem{livezey2021deep}
J.A. Livezey and J.I. Glaser.
\newblock Deep learning approaches for neural decoding: from cnns to lstms and
  spikes to fmri.
\newblock {\em Briefings in Bioinformatics}, 22(2):1577--1591, 2021.

\bibitem{Clark2016surfing}
A.~Clark.
\newblock {\em Surfing Uncertainty: Prediction, Action, and the Embodied Mind}.
\newblock OUP USA, 2016.

\bibitem{Friston2010active}
K.J. Friston, J.~Daunizeau, J.~Kilner, and S.J. Kiebel.
\newblock Action and behavior: a free-energy formulation.
\newblock {\em Biological Cybernetics}, 102(3):227--260, 2010.

\bibitem{Ramstead2018Answering}
D.M.J. Ramstead, P.B. Badcock, and K.J. Friston.
\newblock Answering schrödinger's question: A free-energy formulation.
\newblock {\em Physics of Life Reviews}, 24:1--16, 2018.

\bibitem{Hawkins2004on}
J.~Hawkins and S.~Blakeslee.
\newblock {\em On Intelligence: How a New Understanding of the Brain will Lead
  to the Creation of Truly Intelligent Machines}.
\newblock Times Books, 2004.

\bibitem{Hawkins2019a}
J.~Hawkins, M.~Lewis, M.~Klukas, S.~Purdy, and S.~Ahmad.
\newblock A framework for intelligence and cortical function based on grid
  cells in the neocortex.
\newblock {\em Frontiers in Neural Circuits}, 12:DOI=10.3389/fncir.2018.00121,
  2019.

\bibitem{Hawkins2021a}
J.~Hawkins.
\newblock {\em A Thousand Brains: A New Theory of Intelligence}.
\newblock Basic Books, 2021.

\bibitem{Nagel2001Godel}
E.~Nagel, J.R. Newman, and D.R. Hofstadter.
\newblock {\em Gödel's Proof: Revised Edition}.
\newblock NYU Press, 2001.

\bibitem{Hofstadter1999godel}
D.R. Hofstadter.
\newblock {\em Gödel, Escher, Bach: An Eternal Golden Braid}.
\newblock Basic Books, 1999.

\bibitem{Hofstadter2007i}
D.R. Hofstadter.
\newblock {\em I Am a Strange Loop}.
\newblock Basic Books, 2007.

\bibitem{Hofstadter2013surfaces}
D.R. Hofstadter and E.~Sander.
\newblock {\em Surfaces and Essences: Analogy as the Fuel and Fire of
  Thinking}.
\newblock Basic Books, 2013.

\bibitem{Goodfellow2016deep}
I.~Goodfellow, Y.~Bengio, and A.~Courville.
\newblock {\em Deep Learning: Adaptive Computation and Machine Learning
  series}.
\newblock The MIT Press, 2016.

\bibitem{bengio2021deep}
Y.~Bengio, Y.~Lecun, and Hinton G.
\newblock Deep learning for {AI}.
\newblock {\em Communications of the ACM}, 64(7):58--65, 2021.

\bibitem{Sutton2013reinforcement}
R.S. Sutton and A.G. Barto.
\newblock {\em Reinforcement Learning: An Introduction (second edition)}.
\newblock A Bradford Book, 2018.

\bibitem{Sutton2022the}
R.S. Sutton.
\newblock The quest for a common model of the intelligent decision maker.
\newblock {\em arXiv preprint arXiv:2202.13252v2}, 2022.

\bibitem{wu2019neural}
H.F. Wu, J.Y. Feng, and Y.~Zeng.
\newblock Neural decoding for macaque’s finger position: Convolutional space
  model.
\newblock {\em IEEE Transactions on Neural Systems and Rehabilitation
  Engineering}, 27(3):543--551, 2019.

\bibitem{feng2018neural}
J.Y. Feng, H.F. Wu, and Y.~Zeng.
\newblock Neural decoding for location of macaque's moving finger using
  generative adversarial networks.
\newblock In {\em IEEE International Conference of Intelligent Robotic and
  Control Engineering}, pages 213--217. IEEE, 2018.

\bibitem{feng2020weakly}
J.Y. Feng, H.F. Wu, Y.~Zeng, and Y.H. Wang.
\newblock Weakly supervised learning in neural encoding for the position of the
  moving finger of a macaque.
\newblock {\em Cognitive Computation}, 12(5):1083--1096, 2020.

\bibitem{feng2021vif}
J.Y. Feng, Y.~Luo, and S.~Song.
\newblock {ViF-SD2E}: A robust weakly-supervised method for neural decoding.
\newblock {\em arXiv preprint arXiv:2112.01261v3}, 2021--2022.

\bibitem{Friston2017the}
K.~Friston, T.~Parr, and B.~de~Vries.
\newblock The graphical brain: Belief propagation and active inference.
\newblock {\em Network Neuroscience}, 1(4):381--414, 2017.

\bibitem{Friston2017active}
K.~Friston, T.~FitzGerald, F.~Rigoli, P.~Schwartenbeck, and G.~Pezzulo.
\newblock Active inference: A process theory.
\newblock {\em Neural Computation}, 29(1):1--49, 2017.

\bibitem{Attias2003planning}
H.~Attias.
\newblock Planning by probabilistic inference.
\newblock {\em Proc. of the 9th Int. Workshop on Artificial Intelligence and
  Statistics}, 2003.

\bibitem{Botvinick2012planning}
M.~Botvinick and M.~Toussaint.
\newblock Planning as inference.
\newblock {\em Trends in Cognitive Sciences}, 16(10):485--488, 2012.

\bibitem{Da2020active}
L.~Da~Costa, T.~Parr, N.~Sajid, S.~Veselic, V.~Neacsu, and K.~Friston.
\newblock Active inference on discrete state-spaces: A synthesis.
\newblock {\em Journal of Mathematical Psychology}, 99:102447, 2020.

\bibitem{Lanillos2021active}
P.~Lanillos, C.~Meo, C.~Pezzato, A.A. Meera, M.~Baioumy, W.~Ohata, A.~Tschantz,
  B.~Millidge, M.~Wisse, C.L. Buckley, and J.~Tani.
\newblock Active inference in robotics and artificial agents: Survey and
  challenges.
\newblock {\em arXiv preprint arXiv:2112.01871}, 2021.

\bibitem{Hafting2005Microstructure}
T.~Hafting, M.~Fyhn, S.~Molden, M.B. Moser, and E.I. Moser.
\newblock Microstructure of a spatial map in the entorhinal cortex.
\newblock {\em Nature}, 436(7052):801--806, 2005.

\bibitem{milford2004ratslam}
M.J. Milford, G.F. Wyeth, and D.~Prasser.
\newblock Ratslam: a hippocampal model for simultaneous localization and
  mapping.
\newblock In {\em IEEE International Conference on Robotics and Automation,
  2004. Proceedings. ICRA '04. 2004}, pages 403--408. IEEE, 2004.

\bibitem{yu2019NeuroSLAM}
F.~Yu, J.~Shang, Y.~Hu, and M.~Milford.
\newblock Neuroslam: a brain-inspired slam system for 3d environments.
\newblock {\em Biological Cybernetics}, 113:515--545, 2019.

\bibitem{moser2014grid}
E.I. Moser, Y.~Roudi, M.P. Witter, C.~Kentros, T.~Bonhoeffer, and M.B. Moser.
\newblock Grid cells and cortical representation.
\newblock {\em Nature Reviews Neuroscience}, 15:466--481, 2014.

\bibitem{keefe1971the}
j.~O'Keefe and J.~Dostrovsky.
\newblock The hippocampus as a spatial map. preliminary evidence from unit
  activity in the freely-moving rat.
\newblock {\em Brain Research}, 34(1):171--175, 1971.

\bibitem{taube1990head}
J.S. Taube, R.U. Muller, and J.B.Jr Ranck.
\newblock Head-direction cells recorded from the postsubiculum in freely moving
  rats. i. description and quantitative analysis.
\newblock {\em Journal of Neuroscience}, 10(2):420--435, 1990.

\bibitem{Lever2009boundary}
C.~Lever, S.~Burton, A.~Jeewajee, J.~O'Keefe, and N.~Burgess.
\newblock Boundary vector cells in the subiculum of the hippocampal formation.
\newblock {\em Journal of Neuroscience}, 29(31):9771--9777, 2009.

\bibitem{Barry2007experience}
C.~Barry, R.~Hayman, N.~Burgess, and Jeffery K.J.
\newblock Experience-dependent rescaling of entorhinal grids.
\newblock {\em Nature Neuroscience}, 10:682--684, 2007.

\bibitem{Stensola2012the}
H.~Stensola, T.~Stensola, T.~Solstad, K.~Frøland, M.B. Moser, and Moser E.I.
\newblock The entorhinal grid map is discretized.
\newblock {\em Nature}, 492:72--78, 2012.

\bibitem{Stensola2015shearing}
T.~Stensola, H.~Stensola, M.B. Moser, and Moser E.I.
\newblock Shearing-induced asymmetry in entorhinal grid cells.
\newblock {\em Nature}, 518:207--212, 2015.

\bibitem{Giocomo2015spatial}
L.M. Giocomo.
\newblock Spatial representation: Maps of fragmented space.
\newblock {\em Current Biology}, 25(9):R362--R363, 2015.

\bibitem{Ginosar2021local}
G.~Ginosar, J.~Aljadeff, Y.~Burak, H.~Sompolinsky, L.~Las, and Ulanovsky N.
\newblock Locally ordered representation of 3d space in the entorhinal cortex.
\newblock {\em Nature}, 596:404--409, 2021.

\bibitem{Grieves2021irregular}
R.M. Grieves, S.~Jedidi-Ayoub, K.~Mishchanchuk, A.~Liu, S.~Renaudineau, É.
  Duvelle, and K.J. Jeffery.
\newblock Irregular distribution of grid cell firing fields in rats exploring a
  3d volumetric space.
\newblock {\em Nature Neuroscience}, 24:1567--1573, 2021.

\bibitem{Hohwy2014the}
J.~Hohwy.
\newblock The self-evidencing brain.
\newblock {\em Noûs}, 50(2):259--285, 2014.

\bibitem{Jaynes1957information}
E.T. Jaynes.
\newblock Information theory and statistical mechanics.
\newblock {\em Physical Review Series II}, 106(4):620--630, 1957.

\bibitem{Schmidhuber1991curious}
J.~Schmidhuber.
\newblock Curious model-building control systems.
\newblock {\em In Proc. International Joint Conference on Neural Networks},
  2:1458--1463, 1991.

\bibitem{Schmidhuber2010formal}
J.~Schmidhuber.
\newblock Formal theory of creativity, fun, and intrinsic motivation
  (1990-2010).
\newblock {\em IEEE Transactions on Autonomous Mental Development},
  2(3):230--247, 2010.

\bibitem{Sun2011planning}
Y.~Sun, F.~Gomez, and J.~Schmidhuber.
\newblock Planning to be surprised: optimal bayesian exploration in dynamic
  environments.
\newblock {\em Proceedings of the 4th international conference on Artificial
  general intelligence}, pages 41--51, 2011.

\bibitem{MacKay1995free}
D.J. MacKay.
\newblock Free-energy minimisation algorithm for decoding and cryptoanalysis.
\newblock {\em Electronics Letters}, 31:445--447, 1995.

\bibitem{Wallace1999minimum}
C.S. Wallace and D.L. Dowe.
\newblock Minimum message length and kolmogorov complexity.
\newblock {\em The Computer Journal}, 42(4):270--283, 1999.

\bibitem{Schr1944what}
E.~Schrödinger.
\newblock {\em What is life?}
\newblock Cambridge University Press, 1944.

\bibitem{Godel1931uber}
K.~Gödel.
\newblock Über formal unentscheidbare sätze der principia mathematica und
  verwandter systeme i.
\newblock {\em Monatshefte für Mathematik}, 38(1):173--198, 1931.

\bibitem{Pearl2009causality}
J.~Pearl.
\newblock {\em Causality: Models, Reasoning and Inference}.
\newblock Cambridge University Pressence, 2009.

\bibitem{Pearl2018the}
J.~Pearl and D.~Mackenzie.
\newblock {\em The Book of Why: The New Science of Cause and Effect}.
\newblock Basic Books, 2018.

\bibitem{Liu2021integrated}
Y.~Liu.
\newblock {\em Integrated human-machine intelligence: beyond artificial
  intelligence}.
\newblock Tsing Hua University Press, 2021.

\bibitem{Ha2018world}
D.~Ha and J.~Schmidhuber.
\newblock World models.
\newblock {\em arXiv preprint arXiv:1803.10122v4}, 2018.

\bibitem{Friston2021world}
K.~Friston, R.J. Moran, Y.~Nagai, T.~Taniguchi, H.~Gomi, and J.~Tenenbaum.
\newblock World model learning and inference.
\newblock {\em Neural Networks}, 144:573--590, 2021.

\bibitem{A2022LeCun}
Y.~LeCun.
\newblock A path towards autonomous machine intelligence.
\newblock {\em https://openreview.net/pdf?id=BZ5a1r-kVsf}, 2022.

\bibitem{wang2007introduction}
P.~Wang and B.~Goertzel.
\newblock Introduction: Aspects of artificial general intelligence.
\newblock {\em in Advance of Artificial General Intelligence}, pages 1--16,
  2007.

\bibitem{Zador2023catalyzing}
A.~Zador, S.~Escola, B.~Richards, B.~Ölveczky, Y.~Bengio, K.~Boahen,
  M.~Botvinick, D.~Chklovskii, A.~Churchland, C.~Clopath, J.~DiCarlo,
  S.~Ganguli, J.~Hawkins, K.~Körding, A.~Koulakov, Y.~LeCun, T.~Lillicrap,
  A.~Marblestone, B.~Olshausen, A.~Pouget, C.~Savin, T.~Sejnowski,
  E.~Simoncelli, S.~Solla, D.~Sussillo, A.S. Tolias, and D.~Tsao.
\newblock Catalyzing next-generation artificial intelligence through neuroai.
\newblock {\em nature communications}, 14(1597), 2023.

\bibitem{Chollet2019on}
F.~Chollet.
\newblock On the measure of intelligence.
\newblock {\em arXiv preprint arXiv:1911.01547}, 2019.

\bibitem{hinton2022the}
G.~Hinton.
\newblock The forward-forward algorithm: Some preliminary investigations.
\newblock {\em arXiv preprint arXiv:2212.13345v1}, 2022.

\bibitem{li2023cognitive}
D.Y Li.
\newblock Cognitive physics—the enlightenment by schrödinger, turing, and
  wiener and beyond.
\newblock {\em Intelligent computing}, 2, 2023.

\bibitem{Butlin2023consciousness}
P.~Butlin, R.~Long, E.~Elmoznino, Y.~Bengio, J.~Birch, A.~Constant, G.~Deane,
  S.M. Fleming, C.~Frith, X.~Ji, R.~Kanai, C.~Klein, G.~Lindsay, M.~Michel,
  L.~Mudrik, M.A.K. Peters, E.~Schwitzgebel, J.~Simon, and VanRullen R.
\newblock Consciousness in artificial intelligence: Insights from the science
  of consciousness.
\newblock {\em arXiv preprint arXiv:2308.08708v3}, 2023.

\bibitem{Oizumi2014from}
M.~Oizumi, L.~Albantakis, and G.~Tononi.
\newblock From the phenomenology to the mechanisms of consciousness: Integrated
  information theory 3.0.
\newblock {\em PLOS Computational Biology}, 2014.

\bibitem{Tononi2015Consciousness}
G.~Tononi and C.~Koch.
\newblock Consciousness: here, there and everywhere?
\newblock {\em Philosophical Transactions of the Royal Society B: Biological
  Sciences}, 2015.

\bibitem{Albantakis2021what}
L.~Albantakis and G.~Tononi.
\newblock What we are is more than what we do.
\newblock {\em arXiv preprint arXiv:2102.04219v1}, 2021.

\bibitem{Michel2020on}
M.~Michel and H.~Lau.
\newblock On the dangers of conflating strong and weak versions of a theory of
  consciousness.
\newblock {\em Philosophy and the Mind Sciences}, 2020.

\bibitem{Mediano2022the}
P.A.M. Mediano, F.E. Rosas, D.~Bor, A.K. Seth, and A.B. Barrett.
\newblock The strength of weak integrated information theory.
\newblock {\em Trends in Cognitive Sciences}, 26(8):646--655, 2022.

\bibitem{Bostrom2002anthropic}
Bostrom N.
\newblock {\em Anthropic Bias : Observation Selection Effects in Science and
  Philosophy}.
\newblock Psychology Press, 2002.

\bibitem{Marcus2021insights}
G.~Marcus and E~Davis.
\newblock Insights for ai from the human mind.
\newblock {\em Communications of the ACM}, 64(1):38--41, 2021.

\bibitem{Lenat2023getting}
Lenat D. and G.~Marcus.
\newblock Getting from generative ai to trustworthy ai: What llms might learn
  from cyc.
\newblock {\em arXiv preprint arXiv:2308.04445v1}, 2023.

\bibitem{Hoang2018La}
L.N. Hoang.
\newblock {\em La formule du savoir: Une philosophie unifiée du savoir fondée
  sur le théorème de Bayes}.
\newblock XXX, 2018.

\bibitem{marr2010vision}
D.~Marr and S.~Ullman.
\newblock {\em Vision: A Computational Investigation into the Human
  Representation and Processing of Visual Information}.
\newblock The MIT Press, 2010.

\bibitem{Hatch1993finding}
T.~Hatch and H.~Gardner.
\newblock {\em Finding Cognition in the Classroom: an Expanded View of Human
  Intelligence}.
\newblock New York: Press Syndicate of the University of Cambridge, 1993.

\bibitem{Hutchins1995cognition}
E.~Hutchins.
\newblock {\em Cognition in the Wild}.
\newblock MIT Press, 1995.

\bibitem{Marcus2003the}
G.F. Marcus.
\newblock {\em The Algebraic Mind: Integrating Connectionism and Cognitive
  Science (Learning, Development, and Conceptual Change)}.
\newblock The MIT Press, 2003.

\bibitem{Mitchell2019a}
M.~Mitchell.
\newblock {\em Artificial Intelligence: A Guide for Thinking Humans}.
\newblock Pelican, 2019.

\end{thebibliography}

\begin{thebibliography}{99}
\vspace{3pt}
	\bibitem{1feng2018Neural}
	J.Y. Feng, H.F. Wu, and Y. Zeng.
	\newblock Neural decoding for location of macaque’s moving finger using generative adversarial networks.
	\newblock In {\em IEEE International Conference of Intelligent Robotic and Control Engineering}, pages 213–-217. IEEE, 2018.
	\bibitem{2feng2020weakly}
	J.Y. Feng, H.F. Wu, Y. Zeng, and Y.H. Wang.
	\newblock Weakly supervised learning in neural encoding for the position of the moving finger of a macaque.
	\newblock {\em Cognitive Computation}, 12(5):1083–1096, 2020.
	\bibitem{3feng2021vif}
	J.Y. Feng, Y. Luo, and S. Song.
	\newblock ViF-SD2E: A robust weakly-supervised method for neural decoding.
	\newblock {\em arXiv preprint arXiv:2112.01261v3}, 2021--2022.
    vs. (Final Version). 
	J.Y. Feng, Y. Luo, S. Song, and H. Hu.
	\newblock ViF-SD2E: A robust weakly-supervised framework for neural decoding.
	\newblock {\em Under review}, 2023.
 	\bibitem{4feng2023grid}
	J.Y. Feng, and C.M. Zhang.
	\newblock Grid-SD2E: A General Grid-Feedback in a System for Cognitive Learning.
	\newblock {\em arXiv preprint arXiv:2304.01844v3}, 2023--2024.
	\newblock {\em Under review}, 2023.
    \bibitem{5feng2023grid}
	J.Y. Feng, Y. Luo, and K. Yang.
	\newblock From a symmetry discovered in neural decoding to an algorithm board.
	\newblock {\em Under review}, 2024.

\end{thebibliography}
\renewcommand\refname{$\ast\ast\ast$ (In Figure \ref{fig: Grid-SD2E_board}, from neural coding to cognitive computational learning system) The advances in the study closely related to this work that have been published (2018 -- 2024).}


\newpage
\appendix
\section*{Appendix A}
\label{appendix}
State that before we obtained these formulas, we initially considered factor (1), then progressed to factor (2), and finally to factor (3). We believe that the complexity of Eqs. (1)–(4) obtained using the factor (3) is lower, which does not mean that Eqs. (5)–(8) were incorrect. The calculation formulas for the movement encoder and corrector in Figure \ref{fig: Grid-SD2E_movementspace-2} (b) are given as follows:
\begin{equation}
    {\lim\limits_{{\mathbb R}: N\rightarrow +\infty}} F^{(N)}_{bit, inter}=\left\{
             \begin{array}{llr}
                1\; ;  \quad if \; N=j, \; and, \;  \overline z_k \;or\; z_k \geq f_{mid_1}(\bullet) \\
                0\; ;  \quad if \; N=j, \; and, \;  \overline z_k \;or\; z_k  < f_{mid_1}(\bullet)
             \end{array}
    \right.
\end{equation}
\begin{equation}
    {\lim\limits_{{\mathbb R}: N\rightarrow +\infty}} F^{(N)}_{bit, self}=\left\{
             \begin{array}{llr}
                1\; ;  & if \; N=j, \; and, \;  \overline z_k \;or\; z_k > f_{mid_1}(\bullet) + \epsilon \;\\
                0\; ;  & if \; N=j, \; and, \;  \overline z_k \;or\; z_k < f_{mid_1}(\bullet) - \epsilon \\
                dropout \; or \; F^{(N)}_{bit, inter} & others \\
             \end{array}
    \right.
\end{equation}
\begin{equation}
    {\lim\limits_{{\mathbb R}: N\rightarrow +\infty}} F^{(N)}_{update}=\left\{
             \begin{array}{llr}
                \overline z_{k}\; ;  & if \; N=j, \; and, \; \overline z_{k,bit}=z_{k,bit}  \\
                F^{(N)}_{correct_1}\; ;  & if \; N=j, \; and, \; \overline z_{k,bit} \neq z_{k,bit}
             \end{array}
    \right.
\end{equation}
where $f_{mid_1}(\bullet)$ is the central axis of spatial symmetry at deflection angles $\alpha$ and $\beta$. 
In addition, because the reference centers on the $x'$- and $y'$-axes of a space have different calculation modes, respectively, the predicted values on the $x$- and $y$-axes have different correction formulas, as follows.
\begin{equation}
    F^{(N)}_{correct_1} \approx \left\{
             \begin{array}{lr}
                \overline x_{k} + 2cos^2{\beta}f_x(\eta_1, x_{\max}, x_{\min}, \overline x_{k}) + 2sin{\beta}cos{\beta}f_y(\eta_2, y_{\max}, y_{\min}, \overline y_{k})\; ;  &  if \; N=j, \; \overline z_{k} = \overline x_{k} \\
                \overline y_{k} + 2cos^2{\alpha}f_y(\eta_2, y_{\max}, y_{\min}, \overline y_{k}) - 2sin{\alpha}cos{\alpha}f_x(\eta_1, x_{\max}, x_{\min}, \overline x_{k})\; ;  &  if \; N=j, \; \overline z_{k} = \overline y_{k} 
             \end{array}
    \right.
\end{equation}
where, when $\overline z_{k} = \overline x_{k}$, the calculated deviation of $\overline y_k$ about the $y'$-axis is ignored, and $f_{mid_1}(\bullet) = (\frac{x_{max} + x_{min}}{2}\eta_1 - \overline x_k) cot\beta + \frac{(y_{max} + y_{min})}{2}$, 
$f_x(\eta_1, x_{\max}, x_{\min}, \overline x_k) = \frac{(x_{\max} + x_{\min})}{2}\eta_1 - \overline x_{k}$.
when $\overline z_{k} = \overline y_{k}$, the calculated deviation of $\overline x_k$ about the $x'$-axis is ignored, and $f_{mid_1}(\bullet) = (\overline y_k - \frac{x_{max} + x_{min}}{2}\eta_2) tan\alpha + \frac{(y_{max} + y_{min})}{2}$,
$f_y(\eta_2, y_{\max}, y_{\min}, \overline y_{k}) = (\frac{y_{\max} + y_{\min})}{2}\eta_2 - \overline y_{k}$.
In addition, the deflection angles of the $x$- and $y$-axes have changed, and the origin of the self-built coordinate system is the center of the motion space; then, $\eta_1 = \eta_2 =1$. 
Suppose the deflection angles of the $x$- and $y$-axes have changed, and the origin of the self-built coordinate system is any position of the motion space; then $\eta_1$ and $\eta_2$ are any possible values within a reasonable range.

\section*{Appendix B}
\begin{figure}   
    \centering
    \includegraphics[width=\textwidth]{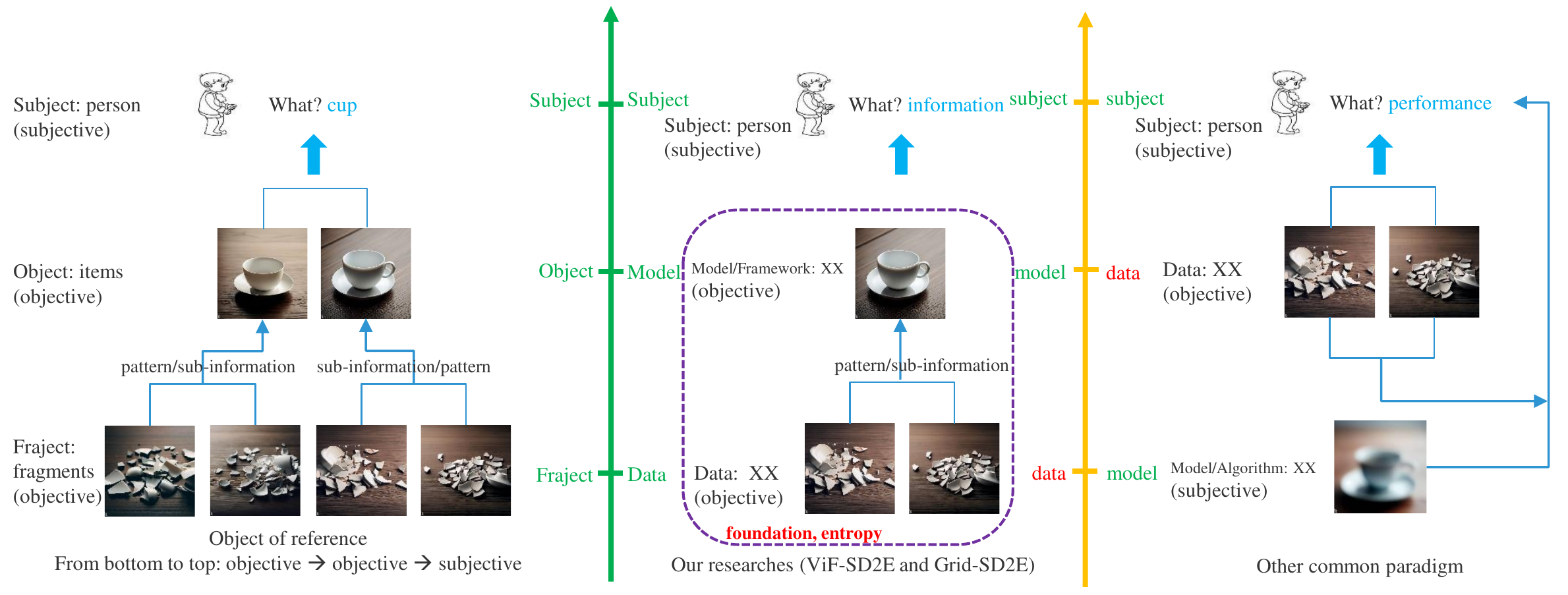}   
    \vspace{-8pt}
    \caption{Two cognitive paradigms of scientific thinking in Appendix B.
    }
    \label{fig: Grid-SD2E_paradigm}
\end{figure}

Object of reference: How do we use the hidden pattern or sub-information of some fragments to put together recognizable objective physical objects and judge that it is a specific cup? (The model and data are closely fitted). Our research: How do we use the hidden pattern or sub-information of some data to build a recognizable objective model and judge that the model contains more information? (The model and data are closely fitted). Other common paradigm: We cannot follow the hidden pattern or sub-information of some data. We have parallel fitting between the data and the model to judge that the model has the best performance. (The model and data are obviously separated).

\newpage
\section*{Appendix C}
\newcommand{\red}[1]{\textcolor{red}{#1}}
\newcommand{\blue}[1]{\textcolor{blue}{#1}}

In addition, the constructed Grid-SD2E is validly verified by data sets from brain data for spatial navigation, but some basic theories still need to be resolved.
Some constructive suggestions have been put forward, and some of them involve more specific research. 
Therefore, these comments are shown as follows for future research and improvement.
Engineering is just one of the researchable topics in the Grid-SD2E framework.
Anonymous comments from others:

1: The manuscript proposed a solution to the well-researched problem of spatial navigation using general learning principles.

2: Feng and Zhang present a machine learning system that incorporates external inputs via a "grid module," Bayesian active inference, and a parameter that tunes the amount of supervision used in the network, or analogously, the balance between exploration and exploitation. Compared to previous work on SD2E, their current work extends upon it by implementing the grid module. This work is compromised by at least four major issues which I will elaborate upon below.

Major issue 1: The authors are presenting a learning paradigm, but they never demonstrated that it could learn anything. That is a critical omission, not to mention the lack of any comparisons with other paradigms. Furthermore, they never showed the internal states of their system to show that they behave as expected. For example, Bayesian updating, exploration vs exploitation, and many other model features can be easily and clearly illustrated for an archetypal task. However, the authors only supported their model with words and architecture diagrams.

Major issue 2: The manuscript suffers from a lack of comprehensibility at both broad and detailed levels. Frequently, variables or terms will be invoked without being clearly described or defined at all. For example, the crucial parameter $N$ is only defined several pages after it was first invoked, and its definition involves "divided spaces," which are not explained. Its purported connection to exploration/exploitation is completely opaque. As another example, the terms deflection angle, processed values, and unprocessed values are used in the caption of Figure 3 completely without definition. The "derivation" in section 2.2.2 makes no sense. What is the movement encoder and corrector? Finally, all of the Figures are inadequately explained by the captions, and some references to them in the text for explanation fail to provide pertinent information. For example, consider the sentence "In addition, as shown in Figures 1 and 4, as the $N$-value increases, the resolution of the positioning continually increases." on page 7. Figures 1 and 4 do not clearly show an increase in positioning resolution with $N$. \textit{\red{Ours: The manuscript has been modified in section \ref{free energy principle}, Figure \ref{fig: Grid-SD2E_movementspace-2}, section \ref{Formula derivation} and section \ref{An introduction}, etc.}}

Major issue 3: The learning system is claimed to relate to the mammalian entorhinal grid system, but no clear relationship can be understood. The closest connection that I can detect is that Grid-SD2E somehow incorporates symmetry upon inversion through the origin; grid cell encodings also contain such points of symmetry. But any more substantial connection eludes me. Where is the 6-fold symmetry? \textit{\red{Ours: The manuscript has been modified in Figure \ref{fig: Grid-SD2E_movementspace} and section \ref{Formula derivation}.}}

Major issue 4: What are the differences between Grid-SD2E and the authors' previous work on SD2E? Of course, the former includes a "grid module" that interacts with the SD module, but what new capabilities does this extension bring? \textit{\red{Ours: The manuscript has been modified in section \ref{introduction}.}} Moreover, they make several lofty and loose claims about connections with Friston's free energy principle, Hofstadter's strange loop theory, and so on. Do these connections only apply to Grid-SD2E or also to SD2E? The connection with Friston's idea, at least, appears trivial; although a very complex argument was delivered, the only significant similarity apparent to me was optimization during learning, which would apply to any learning model.

3: The authors describe a system for cognitive learning, mildly inspired by the grid cells in the context of a system for spatial navigation, but with general learning principles that could potentially be used for various cognitive tasks beyond spatial navigation. They discuss these ideas in relation to previously existing theories from neuroscience, cognitive science, Logic, making it clear that such principles are theoretically needed for such type of cognitive system with general intelligence to be possible. Overally, a lot of details from the proposed system are described and a proper link with previous theory is established but the manuscript text is not very fluid and clear and various spelling and grammar mistakes are in the text. The ideas described could be organized differently in order to make everything clearer and fluid. The following are more specific points to consider:

* Various artificial systems that were partially inspired by grid cells have been proposed such as Rat-SLAM and NeuroSLAM. It would be good if you cite some relevant literature in the introduction, and that would also be useful to contrast the system you are proposing in the sense of describing how it differs from those previous systems. Please describe how it differs (which could also be stated when you mention the main contributions). \textit{\red{Ours: The manuscript has been modified in section \ref{introduction}.}}

* When you first refer to Figure 1 at page 2, please specify where do you overally explain the system depicted in detail. This because it is not clear from Figure 1 what is $S_{k,interaction}$, $S_{k,self}$, $\bar{Z}_k$, $\hat{Z}_k$, etc, please indicate where do you explain that or add the brief explanation either in the caption or when you first refer to the figure. \textit{\red{Ours: The manuscript has been modified in Figure \ref{fig: Grid-SD2E_Procedure}.}}

* Overally, section 2.1 is not very clear mainly regarding:
-The $N$ parameter (please clearly define or explain what does it refer to). \textit{\red{Ours: The manuscript has been modified in section \ref{Distributed hash encoding} and section \ref{free energy principle}.}}
-The symmetry property for the center of activity space (please clearly explain why this is the case for the grid module, citing relevant evidence for it). 

* In section 2.2.2 the explanation for the derivation of formulas (1), (2) and (3) is not very clear. Please use a more clear explanation, referring to the appropriate equation (1), (2) or (3) whenever describing in text. For instance when describing $F^{N}_{bit,inter}$ please refer to equation (1), and so on for the other terms. You mention that you define the function $Z(.)$ but I do not see its equation, please define it and also refer to it whenever relevant. Please explain why do you choose to use dropout for self-reinforcement (equation 2), in the sense of describing what is the underlying probability distribution or probability in that case (also for the case of $F^{N}_{bit,inter}$). \textit{\red{Ours: The manuscript has been modified in section \ref{Formula derivation}.}}

* You also do not explicitly describe how are the signals $\bar{z}_k$, $F^{N}_{bit,inter}$, etc., generated by some underlying neural network system. Please also provide an example or examples of how such signals could be generated by a biological or artificial system, or make it clear to the reader that such details are out of the scope of your proposal, or if they can be generated by some standard techniques/methods. \textit{\red{Ours: The manuscript has been modified in Figure \ref{fig: Grid-SD2E_Procedure} and Figure \ref{fig: Grid-SD2E_movementspace-2}.}}

* This last sentence in section 2.2.2 "Based on the above analysis, the similarity between SD modules and grid cells is shown that corrective (firing) behavior is more prone to occur at nodes in theory" is an important point for your analysis. Please specify what nodes explicitly and elaborate a bit more on this, including how is it that this also happens for grid cells. \textit{\red{Ours: The manuscript has been modified in Figure \ref{fig: Grid-SD2E_movementspace} and section \ref{Formula derivation}.}}

* At section 2.3.1 you mention that the Grid-SD2E system first explores the relevant subspace in an unsupervised way and then selectively switches to supervised learning. Please describe or briefly mention somewhere at the beginning of your manuscript this important feature. \textit{\red{Ours: The manuscript has been modified in section \ref{An introduction}.}}

* At section 2.3.1, equation (5) when you mention about a possible generalization from 2-D to a $d$-dimensional space, you could mention how this might differ from the biological navigational system that uses grid cells. \textit{\red{Ours: The manuscript has been modified in section \ref{An introduction}.}}

* Figure 4 has a nice level of detail but the labels of various variables are too small. Please find a way to zoom-in on the details of Figure 4, either by splitting it or making it bigger. \textit{\red{Ours: The manuscript has been modified in Figure \ref{fig: Grid-SD2E_diagram}.}}

* The relation with the free energy principle and negative entropy in section 2.3.3 is well explained, but please briefly explain equation (6). \textit{\red{Ours: The manuscript has been modified in section \ref{free energy principle}.}}

* In section 3 you mainly describe the connection between the Grid-SD2E system with a special and a general rule for communication between the system and the outside world, including how the system updates its internal structures. You also link such rules with Pearl's causality 3-layer theory for the cognitive ability. I think discussing such connection is the most beneficial point that you contribute in the sense of how from your proposed system can general intelligence possibly emerge. For this, the relation of the complexity of the Grid-SD2E with the free energy principle, which you discuss in section 2.3.3 is also a point you could discuss (in section 4 for instance) together with the special and the general rules. However, the description of ideas in section 3 is not very fluid and clear. For instance, equation (7) is not clearly explained (please explain what do each symbol means exactly and a clear interpretation of it). Also, after equation (7), suddenly you throw the three laws of Human-Machine, which only seem to fill space in your manuscript since you do not make a clear connection between them and the Grid-SD2E system. \textit{\red{Ours: The manuscript has been modified in section \ref{contribution}, section \ref{special and general} and section \ref{paradoxes}.}}

* At section 4 you could elaborate more on the summary of your original contribution. For instance in relation to the two hypotheses that are refined and constitute the Grid-SD2E underlying ideas, you could briefly discuss why is there the hypothesis about the symmetrical attribute and also why for the cognitive system as an incomplete system it requires non-stop interaction (instead of stopping after large enough finite learning time). Please also discuss about the challenges to actually implement your proposed system and also briefly about challenges for generalizing it for other non-spatial domains (in the sense of the link that you establish to general learning theories). \textit{\red{Ours: The manuscript has been modified in section \ref{contribution} and section \ref{challenge}.}}

* At the abstract sentence "... based on the existing theories in both neuroscience and cognitive science,..." I think it would be instructive for the reader if you could very briefly mention what are those theories. \textit{\red{Ours: The manuscript has been modified in section \ref{conclusion}.}}

\end{document}